\documentclass{article}

\usepackage[preprint,nonatbib]{neurips_2024}

\usepackage[utf8]{inputenc} % allow utf-8 input
\usepackage[T1]{fontenc}    % use 8-bit T1 fonts
\usepackage{hyperref}       % hyperlinks
\usepackage{url}            % simple URL typesetting
\usepackage{booktabs}       % professional-quality tables
\usepackage{amsfonts}       % blackboard math symbols
\usepackage{nicefrac}       % compact symbols for 1/2, etc.
\usepackage{microtype}      % microtypography
\usepackage{xcolor}         % colors
\usepackage{amsmath}
\usepackage[linesnumbered,ruled]{algorithm2e} %algorithm
\SetKwInput{kwInit}{Init}
\SetKwInput{KwRequire}{Require}
\usepackage{graphicx}
\usepackage{graphicx, caption, subcaption}
\usepackage{amsmath}

\title{SimSMoE: Solving Representational Collapse via Similarity Measure}

\author{%
Giang Do\thanks{Corresponding author} \quad Hung Le \quad Truyen Tran  \\
Applied Artificial Intelligence Institute (A2I2), Deakin University \\
\texttt{\{s224363215,thai.le,truyen.tran\}@deakin.edu.au}\\
}

\begin{document}

\maketitle

\begin{abstract}
  Sparse mixture of experts (SMoE) have emerged as an effective approach for scaling large language models while keeping a constant computational cost. Regardless of several notable successes of SMoE, effective training such architecture remains elusive due to the representation collapse problem, which in turn harms model performance and causes parameter redundancy. In this work, we present Similarity-based Sparse Mixture of Experts (SimSMoE), a novel similarity of neural network algorithm, that guarantees a solution to address the representation collapse issue between experts given a fixed FLOPs budget. We conduct extensive empirical evaluations on three large language models for both Pre-training and Fine-tuning tasks to illustrate the efficacy, robustness, and scalability of our method. The results demonstrate that SimSMoE significantly enhances existing routing policy and outperforms other SMoE training methods in performance for the tasks.
\end{abstract}

\section{Introduction}

%why do we need SMoE
Large Language Models (LLMs) have demonstrated remarkable achievements across various domains such as natural language processing (NLP) tasks \cite{brown2020language, zhang2022opt, touvron2023llama}, and visual representation learning \cite{jia2021scaling}; \cite{zhu2023minigpt4}. One of the primary reasons for this success is the Transformer \cite{NIPS2017_3f5ee243} architecture and its varied modifications \cite{child2019generating}; \cite{dai2019transformerxl}, which have been scaled up to leverage extensive datasets and advanced computing resources. However, training and inference a single LLM might require hundreds of thousands of compute hours, which is costing millions of dollars \cite{kaddour2023challenges}. This issue motivates contemporary studies to investigate Sparse Mixture-of-Experts (SMoE) \cite{shazeer2017outrageously, zoph2022stmoe, xue2024openmoe, jiang2024mixtral}. SMoE that is inspired by \cite{6797059} usually includes a set of experts with the same architecture and a router network to activate only one or a few experts for each input. Compared to dense models, SMoE reduces inference time thanks to not using all experts simultaneously \cite{artetxe2022efficient, krajewski2024scaling}.

%What is current issue of SMoE
Despite the fact that SMoE has demonstrated its capabilities across various tasks \cite{NEURIPS2021_48237d9f, NEURIPS2022_3e67e84a, gupta2022sparsely}, training efficiency remains a challenge due to the issue of representation collapse, wherein either only a few experts receive routed tokens or all experts converge to learn similar representation. This issue was initially identified and theoretically proven by XMoE \cite{chi2022representation}, followed by consequent works by SMoE-Dropout \cite{chen2023sparse}; HyperRouter \cite{do2023hyperrouter}. To address the limitation, several publications have focused on router policy improvement. Examples include proposals for better routing policies, such as those by Zhou et al.\cite{zhou2022mixtureofexperts}, StableMoE\cite{dai2022stablemoe}, XMoE \cite{chi2022representation}, as well as optimal routing policies like the one suggested by CompeteSMoE \cite{pham2024competesmoe}. These solutions employ indirect approaches that concentrate on token allocation, expecting that enhanced allocation will resolve the collapse among experts. However, the existing methods suffer from several limitations. For example, while XMoE \cite{chi2022representation} and StableMoE \cite{dai2022stablemoe} show promising results, they do not guarantee to solve the representation collapse issue. Additionally, CompeteSMoE \cite{pham2024competesmoe} faces inefficiency problems arising from the requirement to activate all experts. 
%Shortly introduce about the method

In contrast, this paper proposes a novel training framework, named SimSMoE, which directly addresses the collapse issue by emphasizing similar representations among experts. More specifically, we introduce a quantitative method to illustrate the collapse issue between experts using the centered kernel alignment (CKA) metric \cite{kornblith2019similarity}. Then, we see one of the root causes leading to the collapse issue as feeding of the same token between experts. Our effective training strategy comprises three stages: (1) Selecting potential collapsed experts; (2) Identifying collapsed experts; (3) Solving the representation collapse issue. Unlike the router policy approach~\cite{dai2022stablemoe,chi2022representation,do2023hyperrouter}, our framework can be applied to any routing algorithms, as it directly improves expert representations. Moreover, our method guarantees superior SMoE training strategies compared to the existing methods by quantifying the similarity between expert representations and minimizing similarity among experts by the CKA \cite{kornblith2019similarity} loss function. We then evaluate the proposed method by conducting pre-training of Large Language Models (LLMs) on several advanced SMoE architectures, such as GLaM \cite{du2022glam}, Brainformer~\cite{zhou2024brainformers}, or Mistral \cite{jiang2024mixtral}, followed by fine-tuning on downstream tasks. 
%Talk about experiment

%Summary paper contribution
The main contributions of this paper are summarized below.
\begin{itemize}
    \item We demonstrate the representation collapse problem in SMoEs using CKA~\cite{kornblith2019similarity}, which has not been investigated before.
    \item We propose to recognize the collapse among experts and solve the problem by the CKA loss function.
    \item We conduct various experiments on LLMs pre-training and fine-tuning on downstream tasks.
    \item We provide an in-depth analysis of common token feeding and the representation collapse metric, which shows that SimSMoE improves performance compared with existing methods.
\end{itemize}

\section{Background}

\subsection{Sparse Mixture of Experts}

Inspired by conditional computation \cite{NIPS2013_8f1d4362, bengio2013estimating} that activates only some relevant weights of a model on a per-token basis, the Sparse Mixture of Experts (SMoE) model \cite{shazeer2017outrageously}, as an example of conditional computation, with each layer consists \textit{N} experts and a trainable router which selects the most appropriate \textit{k} experts to process  each input sample. 
% In this paper, we apply SMoE for Transformer-based architectures\cite{chi2022representation, dai2022stablemoe, do2023hyperrouter}, and use a  multi-layer perceptron (MLP) as experts architectures, drawing inspiration from \cite{du2022glam, zhou2024brainformers, jiang2024mixtral}.
In this paper, we apply SMoE for Transformer-based architectures\cite{chi2022representation, dai2022stablemoe, do2023hyperrouter} by replacing the feed-forward neural network layer in Transformers\cite{vaswani2023attention} with the Mixture-of-Experts layer, drawing inspiration from \cite{du2022glam, zhou2024brainformers, jiang2024mixtral}.
Each Mixture-of-Experts layer consists of a set of multi-layer perceptrons (MLPs), each with two layers and a ReLu non-linearity function\cite{agarap2019deep}.  
Denoting the output of the multi-head attentions (MHA) as $x$, the output of SMoE with $N$ experts is a weighted sum of each expert's computation $E_i(x)$ by the router function $G(x)$: 
\begin{align} \label{eq:smoe}
    f_{\mathrm{SMoE}}(\boldsymbol{x})=\sum_{i=1}^N G(\boldsymbol{x})_i \cdot E_i(\boldsymbol{x})
    = \sum_{i=1}^N G(\boldsymbol{x})_i \cdot \boldsymbol{W}_{\mathrm{FFN}_i}^2\phi \left(\boldsymbol{W}_{\mathrm{FFN}_i}^1 \boldsymbol{x}\right)
\end{align}

Where $G(x)$ is computed by $TOP_k$ function as equation~(\ref{eq:topk}) that determines the contribution of each expert to the SMoE output.

\begin{align} \label{eq:topk}
    G(\mathbf{x})=\operatorname{TOP}_k(\operatorname{softmax}(\mathbf{W} \mathbf{x}+b))
\end{align}

In this research, we primarily focus on top-2 routing ($K=2$), as studies\cite{NEURIPS2022_2f00ecd7,zoph2022stmoe,sukhbaatar2024branchtrainmix,pham2024competesmoe} have demonstrated its superior balance between training efficiency and testing performance.

\subsection{Challenge of effective Sparse Mixture of Experts Training} \label{sec:collapse}
Recent studies \cite{chi2022representation,do2023hyperrouter} emphasize  the challenge of representation collapse during SMoE training, illustrating that the Jacobian matrix of experts output with respect to input $x \in R^d$ is a linear combination of the expert embeddings ($e \in R^N$). Thus, the phenomenon arises due to $ d >> N$ in practice.  

As the existing solutions \cite{chi2022representation,dai2022stablemoe,do2023hyperrouter,pham2024competesmoe} assume that the collapse problem is a result of ineffective router algorithms, their efforts are directed towards proposing better router mechanisms. 
For a deeper understanding, StableMoE~\cite{dai2022stablemoe} introduces a two-stage training approach, where either routers or experts are exclusively trained in each stage, and XMoE~\cite{chi2022representation} suggests mitigating the collapse problem by employing a deep router featuring down-projection and normalization layers. Additionally, SMoE-Dropout~\cite{chen2023sparse} stabilizes a randomly initialized router and introduces the self-slimmable strategy, gradually expanding the pool of selected experts during training. However, \cite{do2023hyperrouter,pham2024competesmoe} show the solutions still face representation collapse by visualizing router distributions and calculating entropy indexes. Therefore, there is a need to devise a guarantee strategy that focuses on expert representation to alleviate the collapse. 
With this objective in mind, we introduce SimSMoE, presenting two main contributions: (i) Illustrating the collapse problem by a quantitative approach; and (ii) Addressing the issue among experts by CKA loss function~\cite{kornblith2019similarity}.  

\section{Methodology}

We present Similarity Sparse Mixture of Experts (SimSMoE), which utilizes the strengths of existing routing algorithms ~\cite{dai2022stablemoe,chi2022representation,jiang2024mixtral}, directly tackling the representation collapse by minimizing the similarity among expert representations. 

\subsection{SimSMoE} \label{subsec:simsmoe}

\begin{figure*}[t]
    \centering
    \begin{subfigure}{.48\textwidth}
         \centering
         \includegraphics[width=\textwidth]{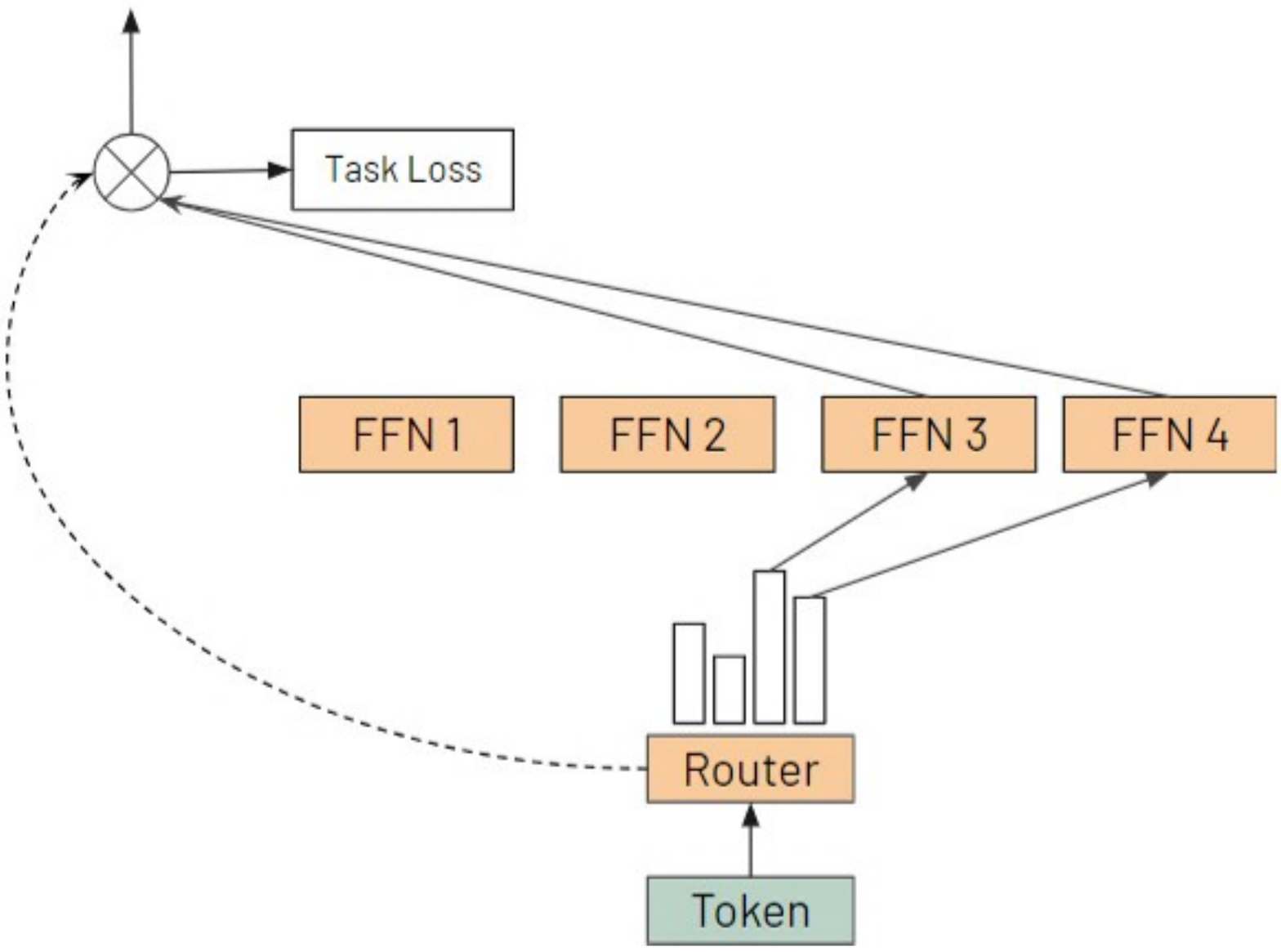}
         \caption{Sparse Mixture-of-Experts (SMoE) Architecture}
         \label{fig:old11}
     \end{subfigure}
     \hfill
    \begin{subfigure}{.48\textwidth}
         \centering
         \includegraphics[width=\textwidth]{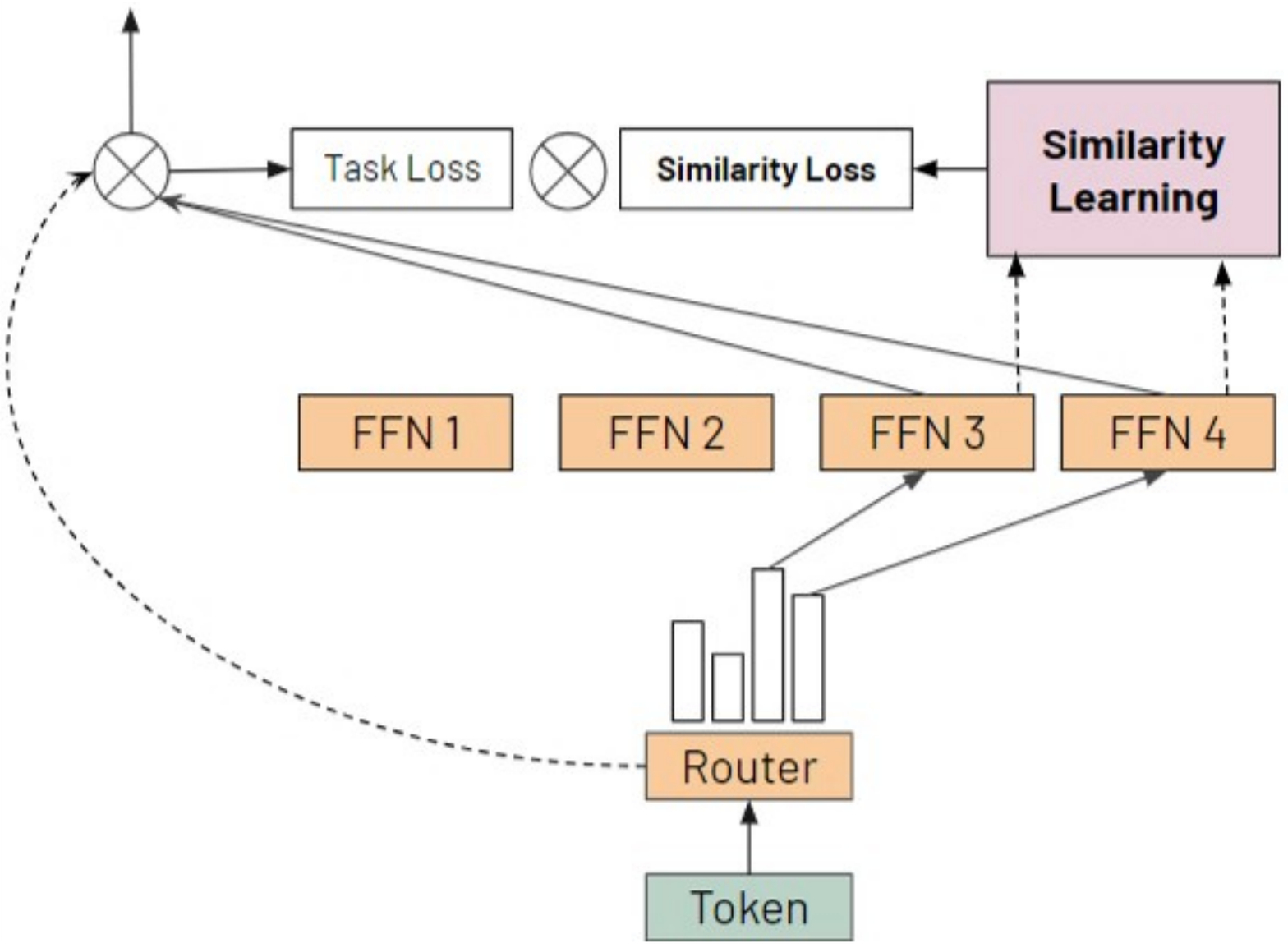}
         \caption{SimSMoE Architecture (Ours)}
         \label{fig:new22}
     \end{subfigure}
     \hfill
     
     \caption{Illustration of the proposed SimSMoE architecture and a SMoE architecture. (a) A SMoE architecture selectively activates experts based on dot-product token-expert routing scores, directing the selected token to the chosen experts. (b) A SimSMoE architecture mitigates the issue of representation collapse by reducing the similarity among the selected experts.} \label{fig:simsmoe}
     \vspace{-0.1in}
\end{figure*}

\textbf{Similarity Reduction.}\; In order to alleviate the representation collapse issue mentioned in Section~\ref{sec:collapse}, we introduce the Similarity Learning module in Figure~\ref{fig:similarity} that helps to minimize the Similarity of Experts Representations. As shown in Figure~\ref{fig:new22}, the Similarity Learning module uses the outputs of experts as input and employs the Similarity Loss described in Section~\ref{sec:simloss} to diversify the experts' representations. 
The key innovation of Similarity Learning consists of two main parts: (i) quantifying  the collapse issue; (ii) diversifying experts' representations using the Similarity Loss described in Section~\ref{sec:simloss}. For more detail, the Similarity Learning is illustrated as Algorithm~\ref{alg:pseudo}. Consequently, the similarity-based SMoE training procedure can be summarized in the following four steps: (1) Calculate the number of shared tokens per expert pair from router $G(x)$, and update the total number of input tokens per expert; (2) Calculate the similarity of selective experts; (3) Update the total loss if the similarity exceeds the similarity threshold; (4) Optimize the total loss in the same manner as training SMoE.

\textbf{An Effective and Reliable Algorithm.}\; One of the biggest challenges for minimizing the similarity among experts is the vast number of possible expert combinations. Given $N$ experts, there are $\binom{N}{2} = \frac{N!}{2! \cdot (N-2)!} = \frac{(N-1) \cdot N}{2} $ expert pairs. To verify the collapse issue of all expert pairs, it is necessary to loop over each pair, calculating their hidden representations and comparing them. This process is equivalent to activating $N$ experts. Due to its contradiction with the conditional computation philosophy of SMoE, proposing an effective algorithm to implement the Similarity Learning is necessary. Section~\ref{subsec:analysis} demonstrates that a higher frequency of common tokens leads to the severity of the collapse. Hence, the training algorithm of SimSMoE introduces two hyperparameters: $f{\ast}$, which represents the frequency for checking the collapse issue in the representation, and $T^{\ast}$, a threshold for identifying the collapse issue as Algorithm~\ref{alg:pseudo}. Indeed, $f{\ast}$ controls computational resources, while $T^{\ast}$ controls the quality of the collapse identification method. Given $T$ as a similarity index between two experts, if $ T \ge T^{\ast}$, it solves the collapse problem. On the other hand, if $ T < T^{\ast}$, the algorithm focuses on optimizing the task loss during the SMoE training process. Thus, if we denote $p$ as the performance of SimSMoE and $p^{\ast}$ as performance of SMoE, we have $ p \ge p^{\ast}$. In addition, both $f{\ast}$ and $T^{\ast}$ are tuned during the training processes.  

The input for the Similarity Learning module comes from a pair of experts. Thus, the most effective way to implement the module is by using the expert outputs from the SMoE training process. The module is workable for top-1 routing, however, it requires activating one additional expert in each iteration. The Similarity Learning module works best for top-k routing ($k \ge 2$), as it fully utilizes the output from pairs of experts to minimize the similarity among them. Additionally, SimSMoE can be applied to any routing algorithm such as StableMoE~\cite{dai2022stablemoe} or XMoE~\cite{chi2022representation} to enhance model performance by addressing the representation collapse problem. We refer to this as a \textit{"Standing on the Shoulders of Giants"} algorithm.

\subsection{Similarity of Neural Network Representations}\label{sec:simloss}

\begin{figure}[t]
    \centering
    \includegraphics[width = 0.6\textwidth]{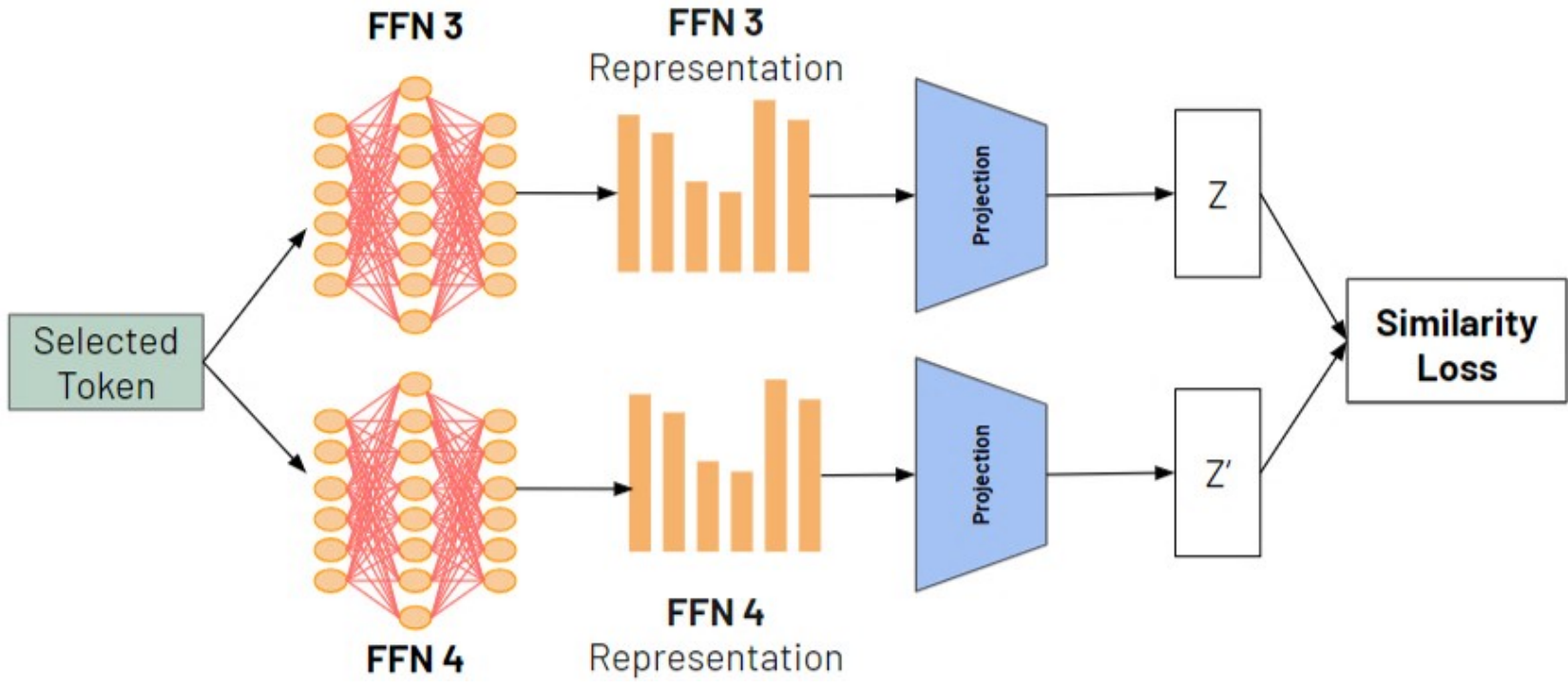}
     \caption{A Similarity Learning Layer (ours) to minimize the similarity among experts.} \label{fig:similarity}
     \vspace{-0.1in}
\end{figure}

Inspired by the Similarity Index~\cite{kornblith2019similarity}, the Similarity Learning module addresses the representation collapse problems from two perspectives. First, the module directly measures a similarity score among experts and helps to identify which experts fail in diversity representation. Then, the Similarity Learning reduces the collapse issue by optimizing the Similarity Loss. Second, the Similarity Learning focuses on solving the collapse at the hidden representations of experts. This allows the method to leverage the advantages of routing techniques such as SMoE with the Balancing Loss~\cite{fedus2022switch}; X-MoE\cite{chi2022representation} StableMoE~\cite{dai2022stablemoe}. Before solving the collapse problem, we first try to identify it, finding that a critical challenge lies in measuring the similarity between neural network representations.
% The Balancing Loss~\cite{fedus2022switch} method uses an auxiliary load balancing loss with a sufficiently high coefficient to ensure effective load balancing among experts. X-MoE~\cite{chi2022representation} estimates the routing scores on a low-dimensional hypersphere by parameterizing the router, where all parameters are learnable, while StableMoE~\cite{dai2022stablemoe} learns a balanced and cohesive routing strategy that is then distilled into a lightweight router that is decoupled from the backbone model. 
To deal with the challenge, Kornblith et al. (2019)~\cite{kornblith2019similarity} proves that the similarity index based on centered kernel alignment (CKA) reliably identifies correspondences between representations in neural networks. In particular, CKA converges to $1.0$ as $ d \to \infty$ that lead to ineffective similarity index. Therefore, drawing inspiration from SimCLR~\cite{chen2020simple} and DirectCLR~\cite{jing2022understanding}, we propose using an MLP with one hidden layer as a projection head (Figure~\ref{fig:similarity}) that maps representations to the space where the similarity loss is applied. Empirically, when scaling the model to larger hidden dimensions, we observe that the projection space can be increased, but one of the good choices is around $N$, with $N$ is number of experts. The centered kernel alignment (CKA) as equation~(\ref{eq:cka}) is a normalized version of Hilbert-Schmidt Independence Criterion (HSIC)~\cite{Gretton2005MeasuringSD} as equation~(\ref{eq:hsic}) that takes values in the interval $[0, 1]$. Kornblith et al. (2019)~\cite{kornblith2019similarity} introduces two versions of CKA: Linear CKA (LCKA) which focuses on linear kernel: $K_{\text {lin }} =\left(x_i \cdot x_j\right)_{i, j}$; and RBF CKA (RCKA) which applies  Gaussian RBF kernel: $K_{G(\sigma)} =\left(e^{\frac{-\left|x_i-x_j\right|^2}{2 \sigma^2}}\right)_{i, j}$. LCKA and RCKA give similar results in practice~\cite{kornblith2019similarity}. For RCKA, selecting bandwidth $\sigma$ determines the extent to which the similarity of small distances is emphasized over large distances. When training the Similarity Learning Layer as Figure~\ref{fig:similarity}, we empirically observe that a larger $\sigma$ performs more stably, so we recommend choosing $\sigma$ in the range of $[0.8, 0.9]$.

\begin{equation} \label{eq:hsic}
\operatorname{HSIC}(K, L)=\frac{1}{(N-1)^2} \operatorname{tr}(KH LH)
\end{equation}

\begin{align} \label{eq:cka}
    \operatorname{CKA}(K, L) = \frac{\operatorname{HSIC}(K, L)} {\sqrt{\operatorname{HSIC}(K, K) \operatorname{HSIC}(L, L)}} = \frac{\operatorname{tr}(K H L H) } {\sqrt{\operatorname{tr}(K H K H) \operatorname{tr}(\text { L H L H })}}
\end{align}
where $\|\|_F$ is the Frobenius norm and $t r$ is the trace function. For RBF CKA, $K$ and $L$ are kernel matrices constructed by evaluating the RBF kernel, and $H$ is the centering matrix $H_n=I_n-\frac{1}{n} \mathbf{1 1}{ }^{\mathrm{T}}$.

\subsection{Training Objective} \label{subsec:training}

The training objective is jointly minimizing the loss of the target task, an auxiliary balancing loss~\cite{fedus2022switch,chi2022representation} ($\mathcal{L}^{\text {balancing }}$) and a similarity loss ($\mathcal{L}^{\text {similarity }}$). Given $K_i$, $L_j$ as the hidden representations of the $i$-th expert and the $j$-th expert respectively, the similarity loss is calculated based on the equation~(\ref{eq:cka}) as follows:

% Given the frequency $f_i$ of how many tokens are routed to the $i$-th expert and the $j$-th expert, the similarity loss is calculated based on equation~(\ref{eq:cka}) as follows:
$$
\mathcal{L}^{\text {similarity }}=  CKA(K_i, L_j)
$$

% $$
% \mathcal{L}^{\text {similarity }}=\frac{N}{|\mathcal{B}|} \cdot \sum_{i=1}^N \sum_{\text {token } \in \mathcal{B}} f_i \cdot CKA(K_i, L_j)
% $$
% where $N$ is the number of the experts, $\mathcal{B}$ is a batch of training examples, $|\mathcal{B}|$ is the number of tokens. 
The overall training objective is to minimize:
$$
\mathcal{L}=\mathcal{L}_{\text {task }}+\alpha \cdot \mathcal{L}^{\text {balancing }} +\beta \cdot \mathcal{L}^{\text {similarity }}
$$
where $\alpha$, $\beta$ are coefficients for the balancing loss and the similarity loss respectively. The term $\mathcal{L}_{\text {task }}$ is defined by the specific task that Large Language Models (LLMs) are learning. For instance, we employ the masked language modeling loss for pre-training and fine-tuning on downstream tasks.

% \begin{algorithm*}[!ht]
% 	\DontPrintSemicolon
% 	\SetKwFunction{algo}{SimSMoE Training}
% 	\SetKwProg{myalg}{Algorithm}{}{}
%         \myalg{\algo{ $\{ t, y_t\}_{i=1}^N$ }}{
% 	\kwRequire {Model architecture $SMoE$, Balancing Loss $B$, Similarity Loss $CKA$, Tracking token per experts $tr$ }
 
% 	\kwInit{Router $R$, $Expert_i$, $Expert_j$, $f{\ast}$, $T^{\ast}$, $\lambda$, $\beta$}

%         \KwResult{$\mathcal{L}$}
        
%         \For{$i \gets 1$ \textbf{to} $N$ }{
%         {Receive a token $t$}

%         {$f_t$ \gets $tr(t)$}

%             \If{$f_t$ \ge $f{\ast}$ }{

%             $\hat{y}_i \vs \gets $  $Expert_i(t)$
            
%             $\hat{y}_j \vs \gets $  $Expert_j(t)$
            
%             $T_t$ \gets $CKA$($\hat{y}_i$, $\hat{y}_j$)
            
%             $\mathcal{L}_{\gR}_{B} \gets \lambda B(R)$
            
%                 \If{$T_t$ \ge $T{\ast}$ }{
                
%                 $\hat{y} \vs \gets $ SMoE($t$) 
                
%                 $\mathcal{L}_{\gR}_{S} \gets \beta T_t$
                
%                 {$\mathcal{L} \gets \gL_{\text{token}}(\hat{y}, y) + \mathcal{L}_{\gR}_{B} + \mathcal{L}_{\gR}$}_{S}   

%                 } 
%                 \Else{
%                 $\hat{y}_t$ \gets SMoE($t$) 

%                 {$\mathcal{L} \gets \gL_{\text{token}}(\hat{y}, y) + \mathcal{L}_{\gR}_{B}$} 
        
%                 }
%             }
%         }

% % \Return $\mathcal{L}$
% }
% \caption{Pseudo-code to train SimSMoE.}
% \label{alg:pseudo}
% \end{algorithm*}

\section{Experiment} \label{sec:exp}

We evaluate SimSMoE on both the mask language modeling task and downstream tasks and compare the performance of the algorithm to other state-of-the-art routing methods for SMoE training. We also present a detailed analysis of the impact of our method in addressing the representation collapse.

\subsection{Experimental Settings}
\textbf{NLP tasks.}\; We investigate two common tasks in pre-training and finetuning of LLMs. Firstly, we perform character-level language modeling on the enwik8~\cite{mahoney_large_2011} or text8 datasets~\cite{mahoney_large_2011}, which are commonly used to evaluate a model’s pre-training capabilities. As is common practice, we follow the default training, validation, and testing splits. Secondly, we finetune the models on downstream applications to investigate their capability to adapt to different domains. For this purpose, we consider pre-trained large models on enwik8 and text8; then finetuning the method on downstream tasks. We select common NLP tasks to evaluate pre-trained models, including the SST-2~\cite{socher_recursive_2013}, SST-5~\cite{socher_recursive_2013}, IMDB~\cite{maas_learning_2011}, and BANKING77~\cite{casanueva-etal-2020-efficient} datasets.

\begin{table*}[!htbp]
\centering
\resizebox{\linewidth}{!}{%
\begin{tabular}{@{}llc|cc|ccc|cc@{}}
\toprule
\multicolumn{2}{c}{Configuration} &\multicolumn{2}{c}{Params}                & \multicolumn{3}{c}{Enwik8 (BPC)} & \multicolumn{3}{c}{Text8 (BPC)}  \\ \midrule
\multicolumn{2}{c}{} &\multicolumn{2}{c}{SimSMoE}                & \multicolumn{2}{c}{SimSMoE} &  & \multicolumn{2}{c}{SimSMoE} & \\ %\midrule
Architecture                 & Algorithm  & No      & Yes  & No            & Yes  & vs. No       & No      & Yes           & vs. No         \\ \midrule
{}  & SMoE     &   &      & 1.11            & \textbf{1.08}   & \textbf{-0.03}       & 1.21       & \textbf{1.20}          & \textbf{-0.01}         \\
{Brainformer}      & XMoE  & 134.4M  & 134.6M        & 1.10            & \textbf{1.09}   & \textbf{-0.01}       & 1.24       & \textbf{1.23}          & \textbf{-0.01}         \\
         & StableMoE  &   &    & 1.10            & \textbf{1.09}   & \textbf{-0.01}       & 1.23       & \textbf{1.20}          & \textbf{-0.03}         \\ \midrule
{}  & SMoE    &  &       & 1.14            & \textbf{1.13}   & \textbf{-0.01}       & 1.26       & \textbf{1.25}          & \textbf{-0.01}         \\
{GLaM}   & XMoE  & 27.8M  & 27.9M       & 1.16            & \textbf{1.14}   & \textbf{-0.02}       & 1.27       & \textbf{1.26}          & \textbf{-0.01}         \\
         & StableMoE   &   &   & 1.16            & \textbf{1.14}   & \textbf{-0.02}       & 1.25       & \textbf{1.24}          & \textbf{-0.01}         \\ \midrule
{}   & SMoE   &   &        & 1.12            & \textbf{1.11}   & \textbf{-0.01}       & 1.23       &    \textbf{1.22}             &       \textbf{-0.01}         \\
{Mistral}   & XMoE    & 62.8M  & 62.9M      & 1.13            & \textbf{1.12}   & \textbf{-0.01}       & 1.24       &       \textbf{1.22}          &       \textbf{-0.02}         \\
    & StableMoE  &   &    & 1.13            & \textbf{1.11}   & \textbf{-0.02}       & 1.23       &    \textbf{1.21}             &     \textbf{-0.02}   
                             \\ \bottomrule
\end{tabular}}
\caption{BPC on the enwik-8 and text8 test sets. Lower is better, best and comparing results are in bold.} \label{table:pre-train}
\end{table*}

\textbf{Architecture.}
We contemplate three advanced SMoE architectures: (i) the Brainformer~\cite{zhou2024brainformers}; (ii) {GLaM}~\cite{du2022glam}; (iii) and Mistral~\cite{jiang2024mixtral}, all of which are decoder-only architectures. Training massive Large Language Models (LLMs) is impractical without substantial industrial resources due to limitations in computational resources. Consequently, we study four model configurations: (i) tiny: with two Brainformer layers and \textbf{3.9M} parameters; (ii) small: with ten GLaM layers and \textbf{27.8M} parameters; and (iii) medium: with seven Mistral layers and \textbf{62.8M} parameters; (iv) large: with ten Brainformer layers and \textbf{134.4M} parameters. 
Rather than striving for state-of-the-art results, we assess the scalability and effectiveness of our algorithm by evaluating multi-scaled models across various datasets.
After that, we run vast investigations using the tiny model to comprehend the behaviors of the algorithm and its robustness to different design choices.

\textbf{Baselines.}
In order to showcase the effectiveness of our method, we establish baselines using the cutting-edge routing methods, including SMoE with the balancing loss~\cite{fedus2022switch}, StableMoE~\cite{dai2022stablemoe}, XMoE~\cite{chi2022representation}. Moreover, these baselines incorporate advanced SMoE architectures such as \textbf{GLaM}~\cite{du2022glam}, \textbf{Brainformer}~\cite{zhou2024brainformers}, \textbf{Mistral}~\cite{jiang2024mixtral}. GLaM\cite{du2022glam} interleaves dense transformer blocks with sparse ones, scaling the capacity of LLMs while significantly reducing training costs compared to dense variants. Brainformer~\cite{zhou2024brainformers}, an improved version of GLaM, further enhances performance by reducing the frequency of attention and modifying layer widths and types, making LLMs faster and more efficient than GLaM. Lastly, Mistral~\cite{jiang2024mixtral} has been successful to scale up LLMs to 34B parameters that outperform the previous state-of-the-art LLMs in reasoning, mathematics, and code generation tasks. \textbf{SMoE} uses a trainable MLP routing mechanism with a \textbf{balancing loss}~\cite{fedus2022switch}, which encourages a balanced load across experts. \textbf{StableMoE}~\cite{dai2022stablemoe} introduces a two-phase training approach, initially focusing solely on training the router and subsequently training the experts with the router fixed, while \textbf{XMoE}~\cite{chi2022representation} features a deep router that includes a down-projection and normalization layer along with a gating network with learnable temperatures.

\textbf{Pre-training and Finetuning.}\; 
SimSMoE fully utilizes all the advantages of routing algorithms, so most of its experimental settings are the same as the baselines for a fair comparison. For the language modeling experiments, we optimize the LLMs pretraining for 50,000 steps using an Adam~\cite{kingma2017adam} optimizer with a linear learning rate schedule. The checkpoint with the lowest validation loss is used to report the final performance on the test set. For routing mechanisms, we apply the default hyper-parameter configurations for both the baselines and SimSMoE. On the top of that, there are two main hyper-parameters only for SimSMoE: the frequency for checking the collapse issue: $f$ and the threshold for identifying the collapse issue: $T$. Next, we cross-validate $f$ with respect to the optimal $T$ found. We use the pretrained checkpoint of Mistral models on enwik8 for each finetuning dataset, and exclude the last layer. Lastly, we employ a randomly initialized fully connected layer as the classifier and finetune all methods for a few epochs using the same learning rate.

\subsection{Language Modeling Evaluation}\label{sec:pretrain}
%%
% \begin{figure}[t]
%     \centering
%     \includegraphics[width = 0.46\textwidth]{valid_plot.pdf}
%      \caption{Validation loss of the small transformer model on enwik8 throughout training.} \label{fig:evolution_valiation}
%      \vspace{-0.2in}
% \end{figure}
%%

\begin{figure*}[t]
    \centering
    \begin{subfigure}{.3\textwidth}
         \centering
         \includegraphics[width=\textwidth]{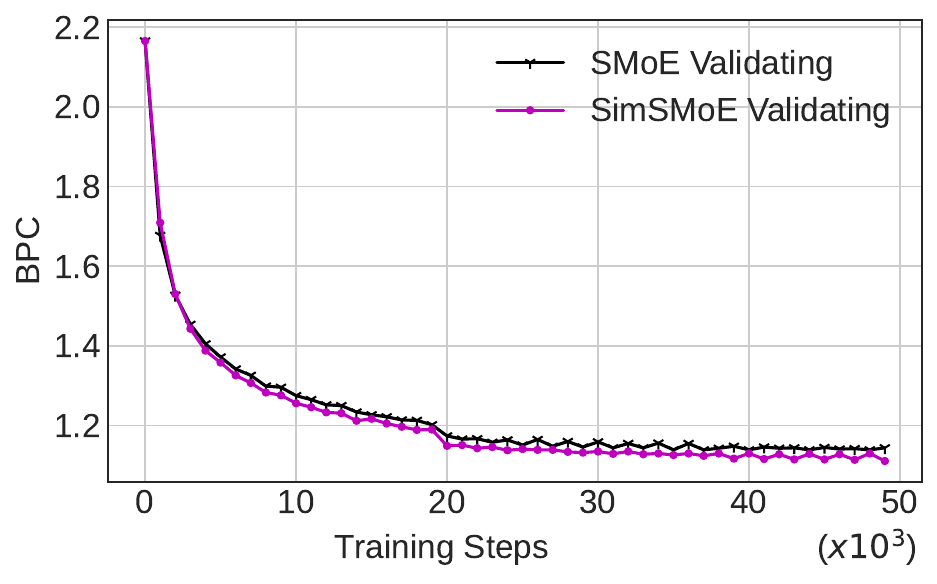}
         \caption{SMoE Routing Policy}
         \label{fig:new}
     \end{subfigure}
     \hfill
     \begin{subfigure}{.3\textwidth}
         \centering
         \includegraphics[width=\textwidth]{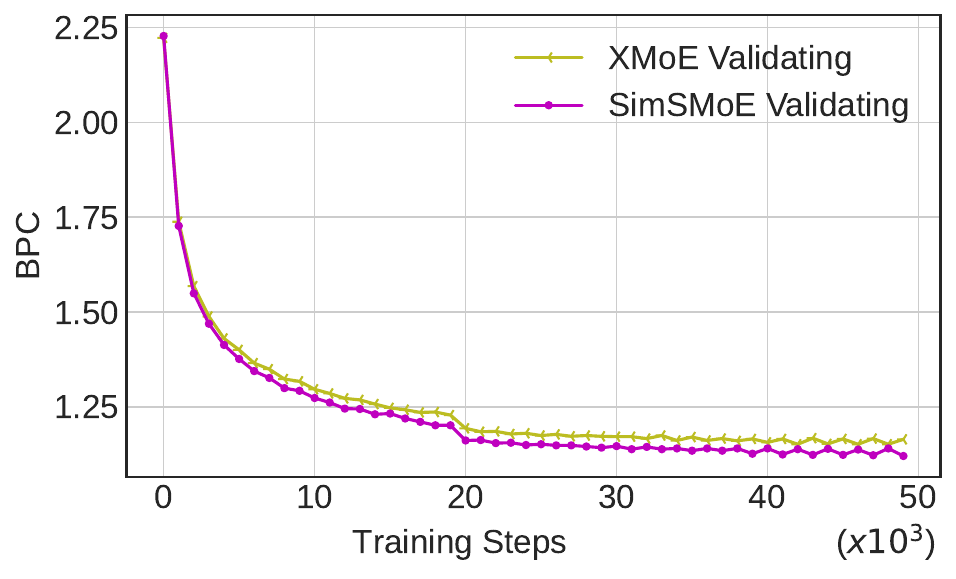}
         \caption{XMoE Routing Policy}
         \label{fig:old33}
     \end{subfigure}
     \hfill
       \begin{subfigure}{.3\textwidth}
         \centering
         \includegraphics[width=\textwidth]{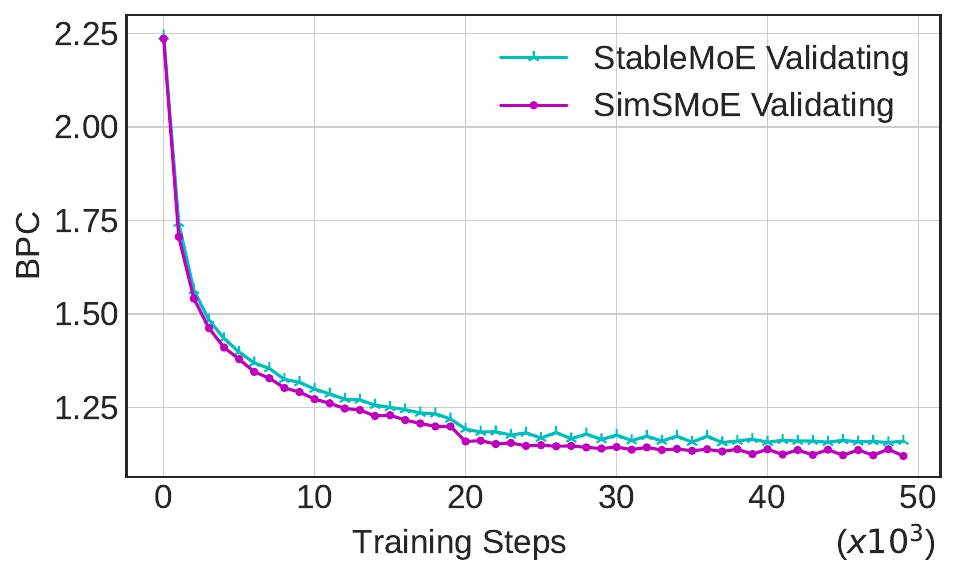}
         \caption{StableMoE Routing Policy}
         \label{fig:old}
     \end{subfigure}
     \hfill
     \caption{Bit-per-Character (BPC) on validation dataset during the training phase reported for Mistral~\cite{jiang2024mixtral} across the three routing mechanisms. (a) SMoE with the Balancing Loss. (b) XMoE. (c) StableMoE} \label{fig:valid_bpc}
     \vspace{-0.1in}
\end{figure*}

In contrast to the baselines, SimSMoE incorporates the Similarity Learning Layer to mitigate representation collapse. As a result, SimSMoE includes an additional \textbf{0.08M} to \textbf{0.16M} parameters compared to the baselines. Table~\ref{table:pre-train} presents the evaluation metrics of SimSMoE versus state-of-the-art strategies. Additionally, we also report the evolution of the performance on the validation set of the SMoE models with various routing policies in Figure~\ref{fig:valid_bpc}.
We initially note that among all routing methods, SimSMoE consistently outperforms the baselines across all datasets for the three decoder-only architectures. Moreover, advanced strategies such as XMoE~\cite{chi2022representation} or StableMoE~\cite{dai2022stablemoe}generally surpass the vanilla SMoE method. Nevertheless, the enhancements achieved by these strategies are often inconsistent or marginal. In contrast, SimSMoE consistently outperforms other competitors on all benchmarks (note that the BPC metric is log-scaled), architectures, and offers a faster convergent rate (Figure~\ref{fig:valid_bpc}). This outcome underscores SimSMoE's proficiency for learning an effective routing policy to facilitate the masked language modeling task.

\subsection{Finetuning Evaluation}
\begin{table}[!ht]
\centering
\resizebox{\linewidth}{!}{%
\begin{tabular}{@{}lc|ccc|ccc|ccc|cc@{}}
\toprule
Method            & \multicolumn{3}{c}{SST-2} & \multicolumn{3}{c}{SST-5} & \multicolumn{3}{c}{IMDB} & \multicolumn{3}{c}{BANKING77} \\ \midrule

            & \multicolumn{2}{c}{SimSMoE} &  & \multicolumn{2}{c}{SimSMoE} & & \multicolumn{2}{c}{SimSMoE} &  & \multicolumn{2}{c}{SimSMoE} & \\ 
Algorithm                & No            & Yes  & vs. No       & No      & Yes           & vs. No  & No      & Yes           & vs. No  & No      & Yes           & vs. No         \\ \midrule

SMoE        & 81.5                      & \textbf{82.8}                      & \textbf{+1.3}                     & 36.9 & \textbf{37.8}  & \textbf{+0.9}  & 85.2  & \textbf{85.7}  & \textbf{+0.5}  & 74.6  & \textbf{79.4}  & \textbf{+4.8}                           \\

XMoE        & 82.2                      & \textbf{82.5}                      & \textbf{+0.3}                     & 34.5    & \textbf{37.4}  & \textbf{+2.9}  & 84.3  & \textbf{84.6}  & \textbf{+0.3}  & 78.6  & \textbf{79.5}  & \textbf{+0.9}                        \\

StableMoE   & 81.0                      & \textbf{82.1}                      & \textbf{+1.1}                     & 36.4  & \textbf{36.7}  & \textbf{+0.3}  & 85.0  & \textbf{85.3}  & \textbf{+0.3}  & 74.1  & \textbf{77.0}  & \textbf{+2.9}                          \\ \bottomrule
\end{tabular} }
\caption{Accuracy of the model after finetuned on various datasets. Higher is better, best and comparing results are in bold.} \label{table:finetune}
\end{table}

Table~\ref{table:finetune} reports the accuracy of the models finetuned on the test sets of various datasets. Overall, we observe that SimSMoE demonstrates strong transfer learning capabilities by achieving the highest accuracy on all datasets. Notably, on the more challenging datasets of SST-5 and BANKING77, which have fewer training samples or more classes, we observe larger performance gains from SimSMoE versus the remaining baselines (over $3\%$ improvements compared to the base methods). This result shows that SimSMoE can boost model performance through solving the collapse issue, which is not only good for pre-training but also exhibits strong transfer capabilities to various downstream tasks. 

\subsection{Ablation Studies} \label{subsec:ablation}
We explore the robustness of SimSMoE under various hyper-parameter settings, conducting all experiments with the tiny Brainformer architecture~\cite{zhou2024brainformers}.

% \begin{table}[h]
   
%     \begin{subtable}[h]{0.25\textwidth}
%         \centering
%         \begin{tabular}{lc}
%         \hline $f^{\ast}$ & BPC \\
%         \hline$1$ & 1.69 \\
%         $4$ & 1.66  \\
%         $8$ & 1.64  \\
%         $16 $ & 1.63  \\
%         \hline SMoE &  1.70 \\ 
%         \hline
%         \end{tabular}
%        \caption{Comparison of frequency of the collapse issue checking for SimSMoE.}
%        \label{tab:table3}
%     \end{subtable}
%     \hfill
%     \begin{subtable}[h]{0.25\textwidth}
%         \centering
%          \begin{tabular}{lc}
%         \hline $T^{\ast}$ & BPC \\
%         \hline$0.1$ & 1.62 \\
%         $0.3$ & 1.63  \\
%         $0.7$ & 1.61  \\
%         $0.9$ & 1.64  \\
%         \hline SMoE &  1.70 \\ 
%         \hline
%         \end{tabular}
%         \caption{Effects of Similarity threshold during pretraining.}
%         \label{tab:table4}
%      \end{subtable}
%      \hfill
%     \begin{subtable}[h]{0.25\textwidth}
%         \centering
%          \begin{tabular}{lc}
%         \hline $\beta$ & BPC \\
%         \hline$0.01$ & 1.61 \\
%         $0.05$ & 1.65  \\
%         $0.1$ & 1.72  \\
%         $0.5$ & 1.75  \\
%         \hline SMoE &  1.70 \\ 
%         \hline
%         \end{tabular}
%         \caption{Pretraining tiny Brainformer results on various Similarity Coefficients.}
%         \label{tab:table5}
%      \end{subtable}
%      \caption{Pretraining tiny Brainformer on enwik8 across different hyperparameter settings}
%      \label{tab:abla}
%     % \end{adjustbox}
% \end{table}

\textbf{SimSMoE Frequency.}\; Since checking the collapse issue for all expert pairs is very costly, as discussed in Section~\ref{subsec:simsmoe}, it is necessary to control computational resources by $f^{\ast}$, which determines the frequency of collapse issue identification. To demonstrate the effectiveness of our algorithm, we analyze the relationship between $f^{\ast}$ and SMoE model performance as the checking frequency increases. All experiments are pretrained under the same settings and evaluated on the enwik8 dataset for a fair comparison. The results reported in Table~\ref{tab:table3} confirm that SimSMoE is effective, consistent with the assumption, as the threshold $f^{\ast}$ increases.

\textbf{Quality Control.}\; In practice, $T^{\ast}$ is a hyperparameter that controls the quality of SimSMoE by determining the level of similarity that can be considered a collapse issue. The value of $T^{\ast}$ ranges from 0 to 1. A low $T^{\ast}$ means more experts pairs are considered collapsed, while a high $T^{\ast}$ means fewer experts are treated as collapsed. Empirically, we find that setting $T^{\ast}$ within the interval $[0.3, 0.7]$ is effective, with a good initial value being 0.5. Table~\ref{tab:table4} shows the pretraining performances of various threshold $T^{\ast}$ on enwik8 dataset.

\textbf{Coefficients of the Similarity Loss.}\; Coefficient $\beta$ determines the weight of the Similarity Loss contribution to the total SMoE Loss. A high value of $\beta$ implies that the model focuses on addressing the collapse, while a low value of $\beta$ indicates the model prioritizes the task loss. Table~\ref{tab:table5} presents the results of the tiny Brainformer across various $\beta$ values.

\subsection{Representation Collapse Analysis} \label{subsec:analysis}

\begin{figure*}[t]
    \centering
    \begin{subfigure}{.48\textwidth}
         \centering
         \includegraphics[width=\textwidth]{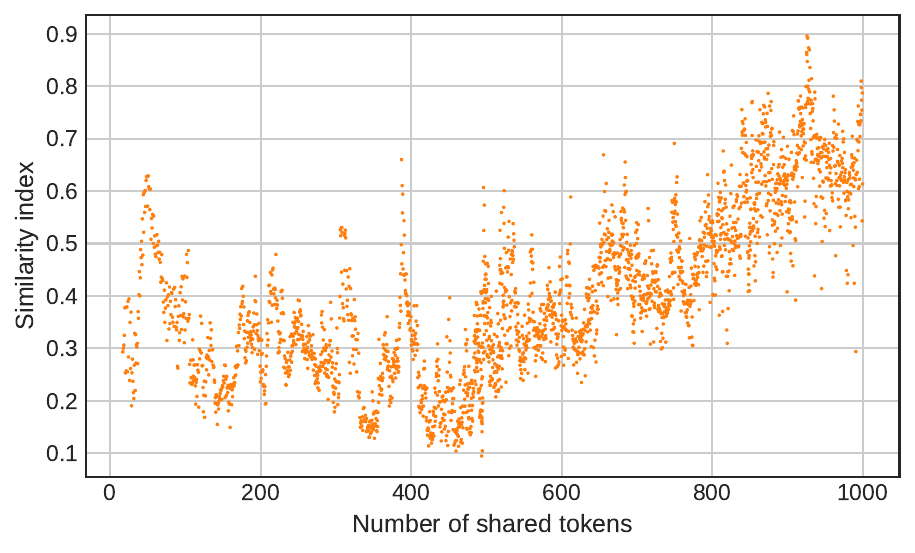}
         \caption{Similarity index.}
         \label{fig:freq}
     \end{subfigure}
     \hfill
    \begin{subfigure}{.48\textwidth}
         \centering
         \includegraphics[width=\textwidth]{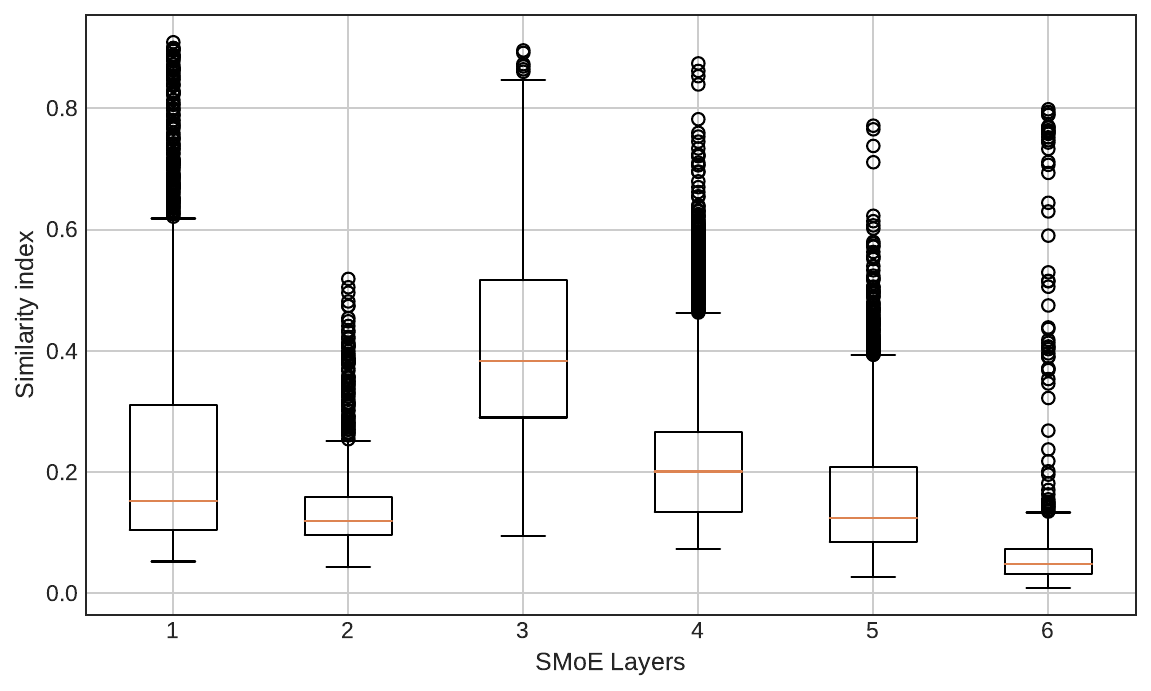}
         \caption{The order of layers.}
         \label{fig:box}
     \end{subfigure}
     \hfill
     
     \caption{Analysis of the similarity index for the Sparse Mixture of Experts (SMoE). Figure (a) shows the correlation between the number of shared tokens and the similarity index. Figure (b) illustrates the similarity index by layer order.} \label{fig:analysis1}
     \vspace{-0.1in}
\end{figure*}

\textbf{Representation  Collapse}\; In a Sparse Mixture of Experts (SMoE) architecture, all experts are typically designed with the same structure, usually as MLPs. To assign tokens to experts, SMoE employs the $TopK$ operator\cite{shazeer2017outrageously}, resulting in certain experts sharing the same tokens. We hypothesize that experts with a high degree of token sharing are more likely to collapse together. To validate our hypothesis, we analyze the correlation between the number of shared tokens and the similarity index among experts. Figure~\ref{fig:analysis1} demonstrates a strong correlation between the number of shared tokens and the similarity among experts, thereby supporting our hypothesis.

\textbf{The order of layers}\; In Section~\ref{subsec:simsmoe}, we discussed that addressing the collapse issue for all pairs of experts is costly. Moreover, since the total loss function described in Section~\ref{subsec:training} is a sum of the task loss, the balancing loss, and the similarity loss, there is a trade-off between resolving the collapse issue and optimizing NLP tasks from a local optimal perspective. Therefore, understanding the differences in collapse levels across layers in SMoE is crucial for effectively allocating resources to address this issue. We visualize the distribution of the similarity index across layers in the Brainformer model, as shown in Figure~\ref{fig:analysis1}. The results indicate that deeper layers exhibit a lower level of collapse compared to earlier layers, suggesting that prioritizing resources to address this issue based on the order of layers in SMoE might improve our method's performance.

\textbf{Similarity Learning Module Effective }\; The Similarity Learning Module is designed to address the issue of representation collapse, which in turn enhances the diversity of the experts' hidden representations. To demonstrate the module's impact, we subtract the hidden representations of two experts in two scenarios: (1) without SimSMoE, and (2) with SimSMoE. Following the suggestions by Samek et al. (2015)~\cite{samek2015evaluating}, we visualize these differences using a heatmap. Without SimSMoE, Figure~\ref{fig:heat} shows how the expert representations become more similar, thus providing support for our method.

\section{Related Work} \label{sec:related}
\subsection{Sparse Mixture of Experts}

\textbf{Sparse Mixture of Experts (SMoE)}\; Motivated by Mixture of Experts (MoE)~\cite{jacobs1991,jordan1994}, Sparse Mixture of Experts (SMoE), with the unifying idea that each example is processed by a subset of the parameters, was first introduced by Shazeer et al. (2017)\cite{shazeer2017outrageously}. SMoE gained further popularity when combined with Transformer large language models~\cite{NEURIPS2022_2f00ecd7,li2022branchtrainmerge,shen2023mixtureofexperts}. After demonstrating promising success in natural language processing, it has been proven in a variety of domains including computer vision~\cite{NEURIPS2021_48237d9f,hwang2023tutel,lin2024moellava}, speech recognition~\cite{wang2023languagerouting,kwon2023}, and multi-task learning~\cite{Ye_2023_ICCV,Chen_2023_CVPR}. However, training SMoE still suffers the representation collapse issue~\cite{chi2022representation}, where all experts converge to similar representation. Chi et al. (2022)~\cite{chi2022representation} identified the issue and proposed XMoE, which estimates the routing scores between tokens and experts on a low-dimensional hypersphere. In subsequent research on the collapse issue, SMoE-dropout\cite{chen2023sparse} suggested that using a randomly initialized and fixed router network to activate experts, and gradually increasing the number of activated experts, might address the problem. Meanwhile, HyperRouter~\cite{do2023hyperrouter} proposed that employing HyperNetwork~\cite{ha2016hypernetworks} to generate router weights is an effective approach for training SMoE. StableMoE~\cite{dai2022stablemoe} also aims to effectively train SMoE by developing a balanced and cohesive routing strategy. This strategy is distilled into a lightweight router, decoupled from the backbone model, which is then used to determine token-to-expert assignments that are frozen to ensure a stable routing strategy. Besides, CompeteSMoE~\cite{pham2024competesmoe} proposed finding an optimal routing using a competition mechanism. Despite tremendous efforts, existing routing solutions have not been able to completely solve the collapse issue~\cite{pham2024competesmoe}. Those methods concentrate on enhancing routing algorithms, whereas our approach is a straightforward solution that directly targets the hidden representation of experts, a topic that remains under-explored.

\subsection{Similarity Learning}

Despite the remarkable empirical advancements of deep neural networks in addressing diverse tasks, understanding and characterizing the representations these networks learn from data remains challenging.
The occurrence of presentation collapse is a common issue in self-supervised learning and has been extensively investigated.~\cite{jing2022understanding,Hua_2021_ICCV,li2022understanding}.
A critical challenge in identifying collapse lies in measuring the similarity between neural network representations. Similarity learning~\cite{kornblith19a,Csisz2021} holds potential for addressing this problem. The current set of representational similarity measures, classified based on their approach to similarity measurement, includes \textit{Canonical Correlation Analysis}~\cite{raghu2017svcca}, \textit{Alignment}~\cite{williams2022generalized}, \textit{Representational Similarity Matrix}~\cite{SHAHBAZI2021118271,10.3389/neuro.06.004.2008}, \textit{Neighbors}~\cite{Wang2023}, \textit{Topology}~\cite{khrulkov2018geometry}, and \textit{Statistic}~\cite{CAMASTRA201626}. Among the aforementioned approaches, the \textit{Representational Similarity Matrix} is widely employed to explore the similarity between the representations of neural networks~\cite{li2016convergent, raghu2017svcca,wang2018understanding,kornblith2019similarity}. Kornblith at el. (2019) emphasized that the canonical correlation analysis (CCA) approach remains invariant under invertible linear transformations only when the retained subspace remains unchanged. They subsequently introduced centered kernel alignment (CKA), which can ascertain the correspondence between the hidden layers of neural networks trained from varying random initializations and widths. Hence, CKA finds extensive application in assessing and alleviating the similarity among neural network representations. For example, One-for-All\cite{hao2023oneforall} utilizes centered kernel alignment (CKA) to compare the learned features between diverse teacher and student models, while Yang at el. (2023)~\cite{NEURIPS2023_1e5fa672} employ CKA to compare across various extractors and hierarchical layers within a single model. In this study, we also illustrate that CKA serves as an appropriate similarity learning metric for addressing representation collapse among experts.

\section{Conclusion and Future Directions}

This study illustrates representation collapse levels in sparse mixture-of-experts (SMoE) models by employing a similarity learning metric. Moreover, we introduce a similarity learning module, which is a direct approach to differentiate expert's hidden representations, designed to alleviate this issue. We proclaim our method as a "Standing on the shoulders of giants" solution, capable of leveraging all the advantages of state-of-the-art routing methods to enhance the efficiency of SMoE training. Furthermore, we believe that directing attention towards expert representation will lead to a promising approach for  training sparse mixtures-of-experts (SMoE) at a large scale. We also extensively evaluate three advanced SMoE architectures for both pre-training and finetuning tasks to demonstrate SimSMoE strong capabilities, scalability, and superiority over state-of-the-art routing strategies. Finally, focusing on expert representation opens up new research avenues for effectively training SMoE, where cutting-edge techniques in Similarity Learning and Contrastive Learning can be harnessed to enhance their performance. We aim to enhance this work from two key perspectives. Firstly, by extending the current SimSMoE approach beyond pairwise expert comparison to leverage entire expert representations, potentially boosting model performance. Secondly, we intend to explore the Contrastive Learning~\cite{chen2020simple} approach as a means to mitigate representation collapse among experts.

\section*{Impact Statements}

\iffalse
This paper presents work whose goal is to advance the field of Machine Learning. There are many potential societal consequences of our work, none which we feel must be specifically highlighted here.
\fi
This paper aims to advance the field of Machine Learning through research conducted in an academic setting using publicly available benchmarks. Our work does not involve human subjects or proprietary assets, and therefore, we believe it has no significant societal implications. However, given that we train large language models on datasets sourced from the web, potential issues such as gender or racial biases may arise, necessitating additional efforts to mitigate their negative impact. Despite the promising results, training a large number of LLMs is inherently expensive and demands substantial computing resources.

\bibliographystyle{plain}
\bibliography{neurips_2024}

\begin{thebibliography}{10}

\bibitem{agarap2019deep}
Abien~Fred Agarap.
\newblock Deep learning using rectified linear units (relu), 2019.

\bibitem{artetxe2022efficient}
Mikel Artetxe, Shruti Bhosale, Naman Goyal, Todor Mihaylov, Myle Ott, Sam Shleifer, Xi~Victoria Lin, Jingfei Du, Srinivasan Iyer, Ramakanth Pasunuru, Giri Anantharaman, Xian Li, Shuohui Chen, Halil Akin, Mandeep Baines, Louis Martin, Xing Zhou, Punit~Singh Koura, Brian O'Horo, Jeff Wang, Luke Zettlemoyer, Mona Diab, Zornitsa Kozareva, and Ves Stoyanov.
\newblock Efficient large scale language modeling with mixtures of experts, 2022.

\bibitem{bengio2013estimating}
Yoshua Bengio, Nicholas Léonard, and Aaron Courville.
\newblock Estimating or propagating gradients through stochastic neurons for conditional computation, 2013.

\bibitem{brown2020language}
Tom~B. Brown, Benjamin Mann, Nick Ryder, Melanie Subbiah, Jared Kaplan, Prafulla Dhariwal, Arvind Neelakantan, Pranav Shyam, Girish Sastry, Amanda Askell, Sandhini Agarwal, Ariel Herbert-Voss, Gretchen Krueger, Tom Henighan, Rewon Child, Aditya Ramesh, Daniel~M. Ziegler, Jeffrey Wu, Clemens Winter, Christopher Hesse, Mark Chen, Eric Sigler, Mateusz Litwin, Scott Gray, Benjamin Chess, Jack Clark, Christopher Berner, Sam McCandlish, Alec Radford, Ilya Sutskever, and Dario Amodei.
\newblock Language models are few-shot learners, 2020.

\bibitem{CAMASTRA201626}
Francesco Camastra and Antonino Staiano.
\newblock Intrinsic dimension estimation: Advances and open problems.
\newblock {\em Information Sciences}, 328:26--41, 2016.

\bibitem{casanueva-etal-2020-efficient}
I{\~n}igo Casanueva, Tadas Tem{\v{c}}inas, Daniela Gerz, Matthew Henderson, and Ivan Vuli{\'c}.
\newblock Efficient intent detection with dual sentence encoders.
\newblock In {\em Proceedings of the 2nd Workshop on Natural Language Processing for Conversational AI}, pages 38--45, Online, July 2020. Association for Computational Linguistics.

\bibitem{chen2023sparse}
Tianlong Chen, Zhenyu Zhang, Ajay Jaiswal, Shiwei Liu, and Zhangyang Wang.
\newblock Sparse moe as the new dropout: Scaling dense and self-slimmable transformers, 2023.

\bibitem{chen2020simple}
Ting Chen, Simon Kornblith, Mohammad Norouzi, and Geoffrey Hinton.
\newblock A simple framework for contrastive learning of visual representations, 2020.

\bibitem{Chen_2023_CVPR}
Zitian Chen, Yikang Shen, Mingyu Ding, Zhenfang Chen, Hengshuang Zhao, Erik~G. Learned-Miller, and Chuang Gan.
\newblock Mod-squad: Designing mixtures of experts as modular multi-task learners.
\newblock In {\em Proceedings of the IEEE/CVF Conference on Computer Vision and Pattern Recognition (CVPR)}, pages 11828--11837, June 2023.

\bibitem{chi2022representation}
Zewen Chi, Li~Dong, Shaohan Huang, Damai Dai, Shuming Ma, Barun Patra, Saksham Singhal, Payal Bajaj, Xia Song, Xian-Ling Mao, Heyan Huang, and Furu Wei.
\newblock On the representation collapse of sparse mixture of experts, 2022.

\bibitem{child2019generating}
Rewon Child, Scott Gray, Alec Radford, and Ilya Sutskever.
\newblock Generating long sequences with sparse transformers, 2019.

\bibitem{Csisz2021}
Adri\'{a}n Csisz\'{a}rik, P\'{e}ter K\H{o}r\"{o}si-Szab\'{o}, \'{A}kos Matszangosz, Gergely Papp, and D\'{a}niel Varga.
\newblock Similarity and matching of neural network representations.
\newblock In M.~Ranzato, A.~Beygelzimer, Y.~Dauphin, P.S. Liang, and J.~Wortman Vaughan, editors, {\em Advances in Neural Information Processing Systems}, volume~34, pages 5656--5668. Curran Associates, Inc., 2021.

\bibitem{dai2022stablemoe}
Damai Dai, Li~Dong, Shuming Ma, Bo~Zheng, Zhifang Sui, Baobao Chang, and Furu Wei.
\newblock Stablemoe: Stable routing strategy for mixture of experts, 2022.

\bibitem{dai2019transformerxl}
Zihang Dai, Zhilin Yang, Yiming Yang, Jaime Carbonell, Quoc~V. Le, and Ruslan Salakhutdinov.
\newblock Transformer-xl: Attentive language models beyond a fixed-length context, 2019.

\bibitem{do2023hyperrouter}
Giang Do, Khiem Le, Quang Pham, TrungTin Nguyen, Thanh-Nam Doan, Bint~T. Nguyen, Chenghao Liu, Savitha Ramasamy, Xiaoli Li, and Steven Hoi.
\newblock Hyperrouter: Towards efficient training and inference of sparse mixture of experts, 2023.

\bibitem{du2022glam}
Nan Du, Yanping Huang, Andrew~M. Dai, Simon Tong, Dmitry Lepikhin, Yuanzhong Xu, Maxim Krikun, Yanqi Zhou, Adams~Wei Yu, Orhan Firat, Barret Zoph, Liam Fedus, Maarten Bosma, Zongwei Zhou, Tao Wang, Yu~Emma Wang, Kellie Webster, Marie Pellat, Kevin Robinson, Kathleen Meier-Hellstern, Toju Duke, Lucas Dixon, Kun Zhang, Quoc~V Le, Yonghui Wu, Zhifeng Chen, and Claire Cui.
\newblock Glam: Efficient scaling of language models with mixture-of-experts, 2022.

\bibitem{fedus2022switch}
William Fedus, Barret Zoph, and Noam Shazeer.
\newblock Switch transformers: Scaling to trillion parameter models with simple and efficient sparsity, 2022.

\bibitem{Gretton2005MeasuringSD}
Arthur Gretton, Olivier Bousquet, Alex Smola, and Bernhard Scholkopf.
\newblock Measuring statistical dependence with hilbert-schmidt norms.
\newblock In {\em International Conference on Algorithmic Learning Theory}, 2005.

\bibitem{gupta2022sparsely}
Shashank Gupta, Subhabrata Mukherjee, Krishan Subudhi, Eduardo Gonzalez, Damien Jose, Ahmed~H. Awadallah, and Jianfeng Gao.
\newblock Sparsely activated mixture-of-experts are robust multi-task learners, 2022.

\bibitem{ha2016hypernetworks}
David Ha, Andrew Dai, and Quoc~V. Le.
\newblock Hypernetworks, 2016.

\bibitem{hao2023oneforall}
Zhiwei Hao, Jianyuan Guo, Kai Han, Yehui Tang, Han Hu, Yunhe Wang, and Chang Xu.
\newblock One-for-all: Bridge the gap between heterogeneous architectures in knowledge distillation, 2023.

\bibitem{Hua_2021_ICCV}
Tianyu Hua, Wenxiao Wang, Zihui Xue, Sucheng Ren, Yue Wang, and Hang Zhao.
\newblock On feature decorrelation in self-supervised learning.
\newblock In {\em Proceedings of the IEEE/CVF International Conference on Computer Vision (ICCV)}, pages 9598--9608, October 2021.

\bibitem{hwang2023tutel}
Changho Hwang, Wei Cui, Yifan Xiong, Ziyue Yang, Ze~Liu, Han Hu, Zilong Wang, Rafael Salas, Jithin Jose, Prabhat Ram, Joe Chau, Peng Cheng, Fan Yang, Mao Yang, and Yongqiang Xiong.
\newblock Tutel: Adaptive mixture-of-experts at scale, 2023.

\bibitem{6797059}
Robert~A. Jacobs, Michael~I. Jordan, Steven~J. Nowlan, and Geoffrey~E. Hinton.
\newblock Adaptive mixtures of local experts.
\newblock {\em Neural Computation}, 3(1):79--87, 1991.

\bibitem{jacobs1991}
Robert~A. Jacobs, Michael~I. Jordan, Steven~J. Nowlan, and Geoffrey~E. Hinton.
\newblock Adaptive mixtures of local experts.
\newblock {\em Neural Computation}, 3(1):79--87, 1991.

\bibitem{jia2021scaling}
Chao Jia, Yinfei Yang, Ye~Xia, Yi-Ting Chen, Zarana Parekh, Hieu Pham, Quoc~V. Le, Yunhsuan Sung, Zhen Li, and Tom Duerig.
\newblock Scaling up visual and vision-language representation learning with noisy text supervision, 2021.

\bibitem{jiang2024mixtral}
Albert~Q. Jiang, Alexandre Sablayrolles, Antoine Roux, Arthur Mensch, Blanche Savary, Chris Bamford, Devendra~Singh Chaplot, Diego de~las Casas, Emma~Bou Hanna, Florian Bressand, Gianna Lengyel, Guillaume Bour, Guillaume Lample, Lélio~Renard Lavaud, Lucile Saulnier, Marie-Anne Lachaux, Pierre Stock, Sandeep Subramanian, Sophia Yang, Szymon Antoniak, Teven~Le Scao, Théophile Gervet, Thibaut Lavril, Thomas Wang, Timothée Lacroix, and William~El Sayed.
\newblock Mixtral of experts, 2024.

\bibitem{jing2022understanding}
Li~Jing, Pascal Vincent, Yann LeCun, and Yuandong Tian.
\newblock Understanding dimensional collapse in contrastive self-supervised learning, 2022.

\bibitem{jordan1994}
Michael Jordan and Robert Jacobs.
\newblock Hierarchical mixtures of experts and the.
\newblock {\em Neural computation}, 6:181--, 01 1994.

\bibitem{kaddour2023challenges}
Jean Kaddour, Joshua Harris, Maximilian Mozes, Herbie Bradley, Roberta Raileanu, and Robert McHardy.
\newblock Challenges and applications of large language models, 2023.

\bibitem{khrulkov2018geometry}
Valentin Khrulkov and Ivan Oseledets.
\newblock Geometry score: A method for comparing generative adversarial networks, 2018.

\bibitem{kingma2017adam}
Diederik~P. Kingma and Jimmy Ba.
\newblock Adam: A method for stochastic optimization, 2017.

\bibitem{kornblith2019similarity}
Simon Kornblith, Mohammad Norouzi, Honglak Lee, and Geoffrey Hinton.
\newblock Similarity of neural network representations revisited, 2019.

\bibitem{kornblith19a}
Simon Kornblith, Mohammad Norouzi, Honglak Lee, and Geoffrey Hinton.
\newblock Similarity of neural network representations revisited.
\newblock In Kamalika Chaudhuri and Ruslan Salakhutdinov, editors, {\em Proceedings of the 36th International Conference on Machine Learning}, volume~97 of {\em Proceedings of Machine Learning Research}, pages 3519--3529. PMLR, 09--15 Jun 2019.

\bibitem{krajewski2024scaling}
Jakub Krajewski, Jan Ludziejewski, Kamil Adamczewski, Maciej Pióro, Michał Krutul, Szymon Antoniak, Kamil Ciebiera, Krystian Król, Tomasz Odrzygóźdź, Piotr Sankowski, Marek Cygan, and Sebastian Jaszczur.
\newblock Scaling laws for fine-grained mixture of experts, 2024.

\bibitem{10.3389/neuro.06.004.2008}
Nikolaus Kriegeskorte, Marieke Mur, and Peter Bandettini.
\newblock Representational similarity analysis - connecting the branches of systems neuroscience.
\newblock {\em Frontiers in Systems Neuroscience}, 2, 2008.

\bibitem{kwon2023}
Yoohwan Kwon and Soo-Whan Chung.
\newblock Mole : Mixture of language experts for multi-lingual automatic speech recognition.
\newblock In {\em ICASSP 2023 - 2023 IEEE International Conference on Acoustics, Speech and Signal Processing (ICASSP)}, pages 1--5, 2023.

\bibitem{li2022understanding}
Alexander~C. Li, Alexei~A. Efros, and Deepak Pathak.
\newblock Understanding collapse in non-contrastive siamese representation learning, 2022.

\bibitem{li2022branchtrainmerge}
Margaret Li, Suchin Gururangan, Tim Dettmers, Mike Lewis, Tim Althoff, Noah~A. Smith, and Luke Zettlemoyer.
\newblock Branch-train-merge: Embarrassingly parallel training of expert language models, 2022.

\bibitem{li2016convergent}
Yixuan Li, Jason Yosinski, Jeff Clune, Hod Lipson, and John Hopcroft.
\newblock Convergent learning: Do different neural networks learn the same representations?, 2016.

\bibitem{lin2024moellava}
Bin Lin, Zhenyu Tang, Yang Ye, Jiaxi Cui, Bin Zhu, Peng Jin, Jinfa Huang, Junwu Zhang, Munan Ning, and Li~Yuan.
\newblock Moe-llava: Mixture of experts for large vision-language models, 2024.

\bibitem{maas_learning_2011}
Andrew~L. Maas, Raymond~E. Daly, Peter~T. Pham, Dan Huang, Andrew~Y. Ng, and Christopher Potts.
\newblock Learning {Word} {Vectors} for {Sentiment} {Analysis}.
\newblock In {\em Proceedings of the 49th {Annual} {Meeting} of the {Association} for {Computational} {Linguistics}: {Human} {Language} {Technologies}}, pages 142--150, Portland, Oregon, USA, June 2011. Association for Computational Linguistics.

\bibitem{mahoney_large_2011}
Matt Mahoney.
\newblock Large text compression benchmark, 2011.

\bibitem{NEURIPS2022_3e67e84a}
Basil Mustafa, Carlos Riquelme, Joan Puigcerver, Rodolphe Jenatton, and Neil Houlsby.
\newblock Multimodal contrastive learning with limoe: the language-image mixture of experts.
\newblock In S.~Koyejo, S.~Mohamed, A.~Agarwal, D.~Belgrave, K.~Cho, and A.~Oh, editors, {\em Advances in Neural Information Processing Systems}, volume~35, pages 9564--9576. Curran Associates, Inc., 2022.

\bibitem{pham2024competesmoe}
Quang Pham, Giang Do, Huy Nguyen, TrungTin Nguyen, Chenghao Liu, Mina Sartipi, Binh~T. Nguyen, Savitha Ramasamy, Xiaoli Li, Steven Hoi, and Nhat Ho.
\newblock Competesmoe -- effective training of sparse mixture of experts via competition, 2024.

\bibitem{raghu2017svcca}
Maithra Raghu, Justin Gilmer, Jason Yosinski, and Jascha Sohl-Dickstein.
\newblock Svcca: Singular vector canonical correlation analysis for deep learning dynamics and interpretability, 2017.

\bibitem{NEURIPS2021_48237d9f}
Carlos Riquelme, Joan Puigcerver, Basil Mustafa, Maxim Neumann, Rodolphe Jenatton, Andr\'{e} Susano~Pinto, Daniel Keysers, and Neil Houlsby.
\newblock Scaling vision with sparse mixture of experts.
\newblock In M.~Ranzato, A.~Beygelzimer, Y.~Dauphin, P.S. Liang, and J.~Wortman Vaughan, editors, {\em Advances in Neural Information Processing Systems}, volume~34, pages 8583--8595. Curran Associates, Inc., 2021.

\bibitem{samek2015evaluating}
Wojciech Samek, Alexander Binder, Grégoire Montavon, Sebastian Bach, and Klaus-Robert Müller.
\newblock Evaluating the visualization of what a deep neural network has learned, 2015.

\bibitem{SHAHBAZI2021118271}
Mahdiyar Shahbazi, Ali Shirali, Hamid Aghajan, and Hamed Nili.
\newblock Using distance on the riemannian manifold to compare representations in brain and in models.
\newblock {\em NeuroImage}, 239:118271, 2021.

\bibitem{shazeer2017outrageously}
Noam Shazeer, Azalia Mirhoseini, Krzysztof Maziarz, Andy Davis, Quoc Le, Geoffrey Hinton, and Jeff Dean.
\newblock Outrageously large neural networks: The sparsely-gated mixture-of-experts layer, 2017.

\bibitem{shen2023mixtureofexperts}
Sheng Shen, Le~Hou, Yanqi Zhou, Nan Du, Shayne Longpre, Jason Wei, Hyung~Won Chung, Barret Zoph, William Fedus, Xinyun Chen, Tu~Vu, Yuexin Wu, Wuyang Chen, Albert Webson, Yunxuan Li, Vincent Zhao, Hongkun Yu, Kurt Keutzer, Trevor Darrell, and Denny Zhou.
\newblock Mixture-of-experts meets instruction tuning:a winning combination for large language models, 2023.

\bibitem{socher_recursive_2013}
Richard Socher, Alex Perelygin, Jean Wu, Jason Chuang, Christopher~D. Manning, Andrew Ng, and Christopher Potts.
\newblock Recursive {Deep} {Models} for {Semantic} {Compositionality} {Over} a {Sentiment} {Treebank}.
\newblock In {\em Proceedings of the 2013 {Conference} on {Empirical} {Methods} in {Natural} {Language} {Processing}}, pages 1631--1642, Seattle, Washington, USA, October 2013. Association for Computational Linguistics.

\bibitem{NIPS2013_8f1d4362}
Rupesh~K Srivastava, Jonathan Masci, Sohrob Kazerounian, Faustino Gomez, and J\"{u}rgen Schmidhuber.
\newblock Compete to compute.
\newblock In C.J. Burges, L.~Bottou, M.~Welling, Z.~Ghahramani, and K.Q. Weinberger, editors, {\em Advances in Neural Information Processing Systems}, volume~26. Curran Associates, Inc., 2013.

\bibitem{sukhbaatar2024branchtrainmix}
Sainbayar Sukhbaatar, Olga Golovneva, Vasu Sharma, Hu~Xu, Xi~Victoria Lin, Baptiste Rozière, Jacob Kahn, Daniel Li, Wen tau Yih, Jason Weston, and Xian Li.
\newblock Branch-train-mix: Mixing expert llms into a mixture-of-experts llm, 2024.

\bibitem{touvron2023llama}
Hugo Touvron, Louis Martin, Kevin Stone, Peter Albert, Amjad Almahairi, Yasmine Babaei, Nikolay Bashlykov, Soumya Batra, Prajjwal Bhargava, Shruti Bhosale, Dan Bikel, Lukas Blecher, Cristian~Canton Ferrer, Moya Chen, Guillem Cucurull, David Esiobu, Jude Fernandes, Jeremy Fu, Wenyin Fu, Brian Fuller, Cynthia Gao, Vedanuj Goswami, Naman Goyal, Anthony Hartshorn, Saghar Hosseini, Rui Hou, Hakan Inan, Marcin Kardas, Viktor Kerkez, Madian Khabsa, Isabel Kloumann, Artem Korenev, Punit~Singh Koura, Marie-Anne Lachaux, Thibaut Lavril, Jenya Lee, Diana Liskovich, Yinghai Lu, Yuning Mao, Xavier Martinet, Todor Mihaylov, Pushkar Mishra, Igor Molybog, Yixin Nie, Andrew Poulton, Jeremy Reizenstein, Rashi Rungta, Kalyan Saladi, Alan Schelten, Ruan Silva, Eric~Michael Smith, Ranjan Subramanian, Xiaoqing~Ellen Tan, Binh Tang, Ross Taylor, Adina Williams, Jian~Xiang Kuan, Puxin Xu, Zheng Yan, Iliyan Zarov, Yuchen Zhang, Angela Fan, Melanie Kambadur, Sharan Narang, Aurelien Rodriguez, Robert Stojnic, Sergey Edunov, and Thomas
  Scialom.
\newblock Llama 2: Open foundation and fine-tuned chat models, 2023.

\bibitem{NIPS2017_3f5ee243}
Ashish Vaswani, Noam Shazeer, Niki Parmar, Jakob Uszkoreit, Llion Jones, Aidan~N Gomez, \L~ukasz Kaiser, and Illia Polosukhin.
\newblock Attention is all you need.
\newblock In I.~Guyon, U.~Von Luxburg, S.~Bengio, H.~Wallach, R.~Fergus, S.~Vishwanathan, and R.~Garnett, editors, {\em Advances in Neural Information Processing Systems}, volume~30. Curran Associates, Inc., 2017.

\bibitem{vaswani2023attention}
Ashish Vaswani, Noam Shazeer, Niki Parmar, Jakob Uszkoreit, Llion Jones, Aidan~N. Gomez, Lukasz Kaiser, and Illia Polosukhin.
\newblock Attention is all you need, 2023.

\bibitem{Wang2023}
Chenxu Wang, Wei Rao, Wenna Guo, Pinghui Wang, Jun Liu, and Xiaohong Guan.
\newblock Towards understanding the instability of network embedding (extended abstract).
\newblock In {\em 2023 IEEE 39th International Conference on Data Engineering (ICDE)}, pages 3825--3826, 2023.

\bibitem{wang2018understanding}
Liwei Wang, Lunjia Hu, Jiayuan Gu, Yue Wu, Zhiqiang Hu, Kun He, and John Hopcroft.
\newblock Towards understanding learning representations: To what extent do different neural networks learn the same representation, 2018.

\bibitem{wang2023languagerouting}
Wenxuan Wang, Guodong Ma, Yuke Li, and Binbin Du.
\newblock Language-routing mixture of experts for multilingual and code-switching speech recognition, 2023.

\bibitem{williams2022generalized}
Alex~H. Williams, Erin Kunz, Simon Kornblith, and Scott~W. Linderman.
\newblock Generalized shape metrics on neural representations, 2022.

\bibitem{xue2024openmoe}
Fuzhao Xue, Zian Zheng, Yao Fu, Jinjie Ni, Zangwei Zheng, Wangchunshu Zhou, and Yang You.
\newblock Openmoe: An early effort on open mixture-of-experts language models, 2024.

\bibitem{NEURIPS2023_1e5fa672}
mengping yang, Ceyuan Yang, Yichi Zhang, Qingyan Bai, Yujun Shen, and Bo~Dai.
\newblock Revisiting the evaluation of image synthesis with gans.
\newblock In A.~Oh, T.~Naumann, A.~Globerson, K.~Saenko, M.~Hardt, and S.~Levine, editors, {\em Advances in Neural Information Processing Systems}, volume~36, pages 9518--9542. Curran Associates, Inc., 2023.

\bibitem{Ye_2023_ICCV}
Hanrong Ye and Dan Xu.
\newblock Taskexpert: Dynamically assembling multi-task representations with memorial mixture-of-experts.
\newblock In {\em Proceedings of the IEEE/CVF International Conference on Computer Vision (ICCV)}, pages 21828--21837, October 2023.

\bibitem{zhang2022opt}
Susan Zhang, Stephen Roller, Naman Goyal, Mikel Artetxe, Moya Chen, Shuohui Chen, Christopher Dewan, Mona Diab, Xian Li, Xi~Victoria Lin, Todor Mihaylov, Myle Ott, Sam Shleifer, Kurt Shuster, Daniel Simig, Punit~Singh Koura, Anjali Sridhar, Tianlu Wang, and Luke Zettlemoyer.
\newblock Opt: Open pre-trained transformer language models, 2022.

\bibitem{zhou2024brainformers}
Yanqi Zhou, Nan Du, Yanping Huang, Daiyi Peng, Chang Lan, Da~Huang, Siamak Shakeri, David So, Andrew Dai, Yifeng Lu, Zhifeng Chen, Quoc Le, Claire Cui, James Laudon, and Jeff Dean.
\newblock Brainformers: Trading simplicity for efficiency, 2024.

\bibitem{zhou2022mixtureofexperts}
Yanqi Zhou, Tao Lei, Hanxiao Liu, Nan Du, Yanping Huang, Vincent Zhao, Andrew Dai, Zhifeng Chen, Quoc Le, and James Laudon.
\newblock Mixture-of-experts with expert choice routing, 2022.

\bibitem{NEURIPS2022_2f00ecd7}
Yanqi Zhou, Tao Lei, Hanxiao Liu, Nan Du, Yanping Huang, Vincent Zhao, Andrew~M Dai, zhifeng Chen, Quoc~V Le, and James Laudon.
\newblock Mixture-of-experts with expert choice routing.
\newblock In S.~Koyejo, S.~Mohamed, A.~Agarwal, D.~Belgrave, K.~Cho, and A.~Oh, editors, {\em Advances in Neural Information Processing Systems}, volume~35, pages 7103--7114. Curran Associates, Inc., 2022.

\bibitem{zhu2023minigpt4}
Deyao Zhu, Jun Chen, Xiaoqian Shen, Xiang Li, and Mohamed Elhoseiny.
\newblock Minigpt-4: Enhancing vision-language understanding with advanced large language models, 2023.

\bibitem{zoph2022stmoe}
Barret Zoph, Irwan Bello, Sameer Kumar, Nan Du, Yanping Huang, Jeff Dean, Noam Shazeer, and William Fedus.
\newblock St-moe: Designing stable and transferable sparse expert models, 2022.

\end{thebibliography}

%%%%%%%%%%%%%%%%%%%%%%%%%%%%%%%%%%%%%%%%%%%%%%%%%%%%%%%%%%%%%%%%%%%%%%%%%%%%%%%
%%%%%%%%%%%%%%%%%%%%%%%%%%%%%%%%%%%%%%%%%%%%%%%%%%%%%%%%%%%%%%%%%%%%%%%%%%%%%%%
% APPENDIX
%%%%%%%%%%%%%%%%%%%%%%%%%%%%%%%%%%%%%%%%%%%%%%%%%%%%%%%%%%%%%%%%%%%%%%%%%%%%%%%
%%%%%%%%%%%%%%%%%%%%%%%%%%%%%%%%%%%%%%%%%%%%%%%%%%%%%%%%%%%%%%%%%%%%%%%%%%%%%%%
\newpage
\appendix
\onecolumn

\begin{center}
{\bf{\Large{Supplementary Material for ``SimSMoE: Solving Representational Collapse via Similarity Measure"}}}
\end{center}

This document is structured as follows: Appendix~\ref{app:add1} provides detail materials for SimSMoE algorithm, ablation studies results, and representation collapse analysis. Appendix~\ref{app:add_exp2} offers a detailed settings for our experiments in Section~\ref{sec:exp}.

% Appendix~\ref{app:check} contains all materials for the requirement checklists. 

\section{Additional Materials} \label{app:add1}

\subsection{SimSMoE Algorithm details} \label{app:add_exp3}
The training procedure for similarity-based SMoE can be succinctly outlined in four steps. First, compute the shared tokens per expert pair through router $G(x)$, updating the total input tokens for each expert accordingly to verify the frequency condition. Next, assess the similarity among chosen experts. If this similarity surpasses the predefined threshold, proceed to update the total loss. Finally, refine the overall loss using the same optimization approach employed in traditional SMoE training.

% \begin{algorithm*}[!ht]
% 	\DontPrintSemicolon
% 	\SetKwFunction{algo}{SimSMoE Training}
% 	\SetKwProg{myalg}{Algorithm}{}{}
%         \myalg{\algo{ $\{ t, y_t\}_{i=1}^N$ }}{
% 	\KwRequire {Model architecture $SMoE$, Balancing Loss $B$, Similarity Loss $CKA$, Tracking token per experts $tr$ }
 
% 	\kwInit{Router $R$, $Expert_i$, $Expert_j$, $f{\ast}$, $T^{\ast}$, $\lambda$, $\beta$}

%         \KwResult{$\mathcal{L}$}
        
%         \For{$i \gets 1$ \textbf{to} $N$ }{
%         {Receive a token $t$}

%         {$f_t$ \gets $tr(t)$}

%             \If{$f_t$ \ge $f{\ast}$ }{

%             $\hat{y}_i \vs \gets $  $Expert_i(t)$
            
%             $\hat{y}_j \vs \gets $  $Expert_j(t)$
            
%             $T_t$ \gets $CKA$($\hat{y}_i$, $\hat{y}_j$)
            
%             $\mathcal{L}_{\gR}_{B} \gets \lambda B(R)$
            
%                 \If{$T_t$ \ge $T{\ast}$ }{
                
%                 $\hat{y} \vs \gets $ SMoE($t$) 
                
%                 $\mathcal{L}_{\gR}_{S} \gets \beta T_t$
                
%                 {$\mathcal{L} \gets \gL_{\text{token}}(\hat{y}, y) + \mathcal{L}_{\gR}_{B} + \mathcal{L}_{\gR}$}_{S}   

%                 } 
%                 \Else{
%                 $\hat{y}_t$ \gets SMoE($t$) 

%                 {$\mathcal{L} \gets \gL_{\text{token}}(\hat{y}, y) + \mathcal{L}_{\gR}_{B}$} 
        
%                 }
%             }
%         }

% % \Return $\mathcal{L}$
% }
% \caption{Pseudo-code to train SimSMoE.}
% \label{alg:pseudo}
% \end{algorithm*}

\begin{algorithm*}[!ht]
\DontPrintSemicolon
\SetKwFunction{algo}{SimSMoE Training}
\SetKwProg{myalg}{Algorithm}{}{}
\myalg{\algo{$\{ t, y_t\}_{i=1}^N$}}{
    \KwRequire{Model architecture $SMoE$, Balancing Loss $B$, Similarity Loss $CKA$, Tracking token per experts $tr$} \;
    \kwInit{Router $R$, $Expert_i$, $Expert_j$, $f{\ast}$, $T^{\ast}$, $\lambda$, $\beta$} \;
    \KwResult{$\mathcal{L}$} \;
    
    \For{$i \gets 1$ \textbf{to} $N$}{
        Receive a token $t$ \;
        $f_t \gets tr(t)$ \;
        
        \If{$f_t \ge f^{\ast}$}{
            $\hat{y}_i \gets Expert_i(t)$ \;
            $\hat{y}_j \gets Expert_j(t)$ \;
            $T_t \gets CKA(\hat{y}_i, \hat{y}_j)$ \;
            $\mathcal{L}_{B} \gets \lambda B(R)$ \;
            
            \If{$T_t \ge T^{\ast}$}{
                $\hat{y} \gets SMoE(t)$ \;
                $\mathcal{L}_{S} \gets \beta T_t$ \;
                $\mathcal{L} \gets \mathcal{L}_{\text{token}}(\hat{y}, y) + \mathcal{L}_{B} + \mathcal{L}_{S}$ \;
            } 
            \Else{
                $\hat{y}_t \gets SMoE(t)$ \;
                $\mathcal{L} \gets \mathcal{L}_{\text{token}}(\hat{y}, y) + \mathcal{L}_{B}$ \;
            }
        }
    }
}
\caption{Pseudo-code to train SimSMoE.}
\label{alg:pseudo}
\end{algorithm*}

\subsection{Ablation Studies results} \label{app:add_exp}

\begin{table}[!ht]
   
    \begin{subtable}[h]{0.25\textwidth}
        \centering
        \begin{tabular}{lc}
        \hline $f^{\ast}$ & BPC \\
        \hline$1$ & 1.56 \\
        $4$ & 1.58  \\
        $8$ & 1.55  \\
        $16 $ & \textbf{1.54}  \\
        \hline SMoE &  1.69 \\ 
        \hline
        \end{tabular}
       \caption{Comparison of frequency of the collapse issue checking for SimSMoE.}
       \label{tab:table3}
    \end{subtable}
    \hfill
    \begin{subtable}[h]{0.25\textwidth}
        \centering
         \begin{tabular}{lc}
        \hline $T^{\ast}$ & BPC \\
        \hline$0.1$ & 1.54 \\
        $0.3$ & 1.55  \\
        $0.3$ & \textbf{1.54}  \\
        $0.7$ & 1.55  \\
        $0.9$ & 1.55  \\
        \hline SMoE &  1.69 \\ 
        \hline
        \end{tabular}
        \caption{Effects of Similarity threshold during pretraining.}
        \label{tab:table4}
     \end{subtable}
     \hfill
    \begin{subtable}[h]{0.25\textwidth}
        \centering
         \begin{tabular}{lc}
        \hline $\beta$ & BPC \\
        \hline$0.005$ & 1.55 \\
        $0.01$ & \textbf{1.54} \\
        $0.05$ & 1.56  \\
        $0.1$ & 1.54  \\
        $0.2$ & 1.57  \\
        \hline SMoE &  1.69 \\ 
        \hline
        \end{tabular}
        \caption{Pretraining tiny Brainformer results on various Similarity Coefficients.}
        \label{tab:table5}
     \end{subtable}
     \caption{Pretraining tiny Brainformer on enwik8 across different hyperparameter settings}
     \label{tab:abla}
    % \end{adjustbox}
\end{table}

\subsection{Representation Collapse Analysis}

\begin{figure*}[t]
    \centering
    \begin{subfigure}{.48\textwidth}
         \centering
         \includegraphics[width=\textwidth]{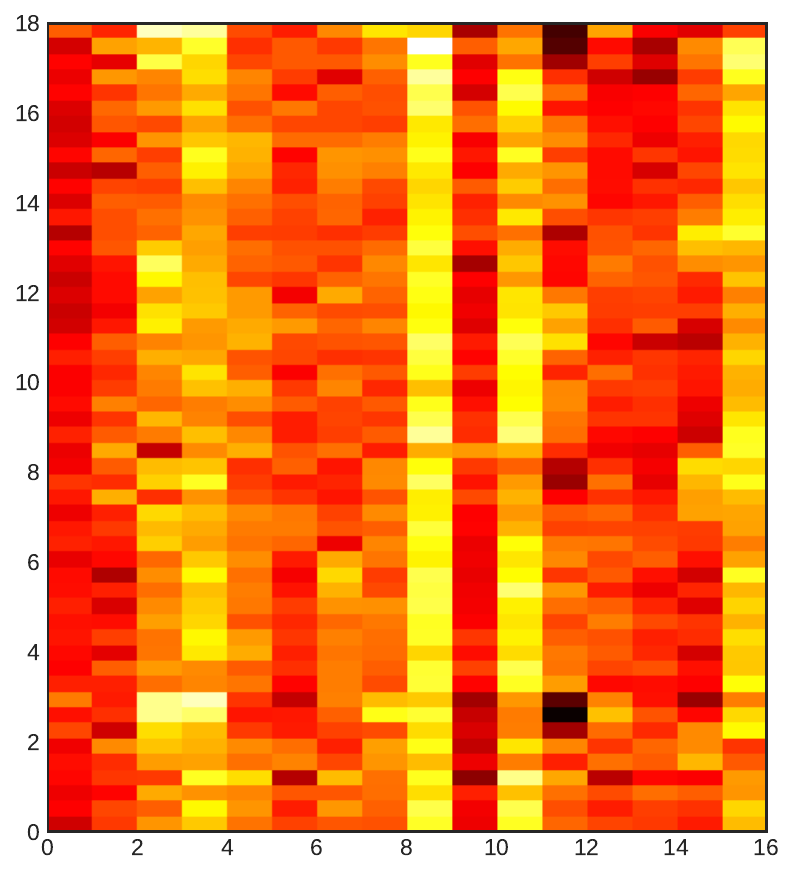}
         \caption{SMoE Layer.}
         \label{fig:sm}
     \end{subfigure}
     \hfill
    \begin{subfigure}{.48\textwidth}
         \centering
         \includegraphics[width=\textwidth]{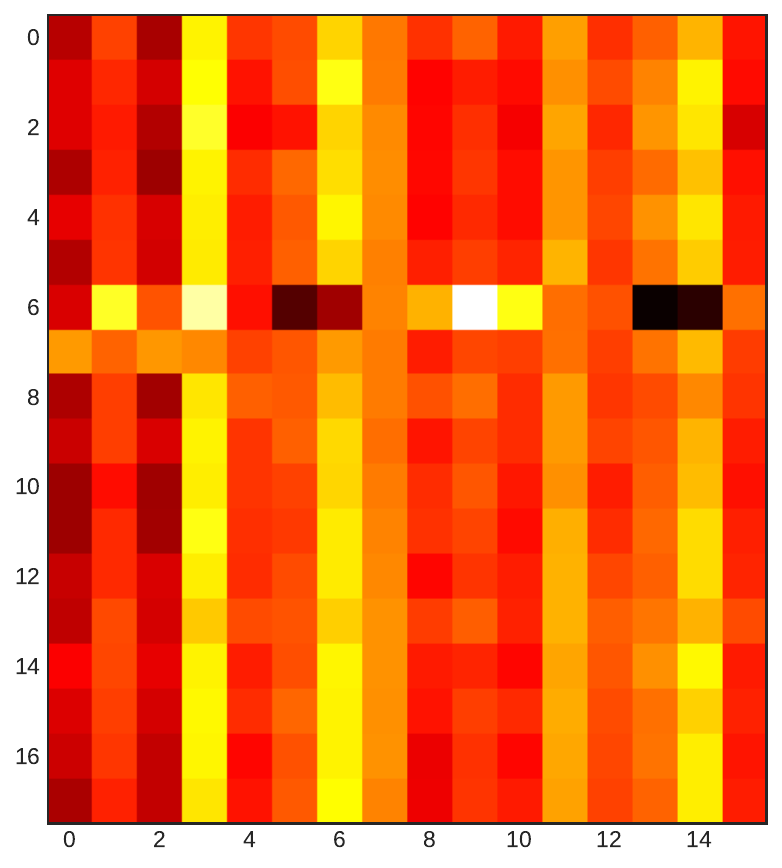}
         \caption{SimSMoE Layer.}
         \label{fig:ss}
     \end{subfigure}
     \hfill
     
     \caption{Exploration of the impact of similarity learning on diversity model representation. Figure (a) shows the heatmap of differences between the hidden representations of two experts for the SMoE layer. Figure (b) shows the heatmap of differences between the hidden representations of two experts for the SimSMoE layer.} \label{fig:heat}
     \vspace{-0.1in}
\end{figure*}

\section{Experiments implementation details} \label{app:add_exp2}

This section provides detailed parameters of our experiments in Section \ref{sec:exp}. 

\subsection{General Settings} \label{app:setting}
% The experiments are developed based on the publicly available CompeteSMoE \citep{pham2024competesmoe} implementation\footnote{\url{https://github.com/giangdip2410/CompeteSMoE}}. However, the pre-training models were conducted on a single A100 GPU. Therefore, it might yield different results when parallel training on multiple GPUs. 
The experiments are based on the publicly available CompeteSMoE implementation\cite{pham2024competesmoe}\footnote{\url{https://github.com/giangdip2410/CompeteSMoE}}. However, the pre-training was conducted on a single A100 GPU, so results might differ when using parallel training on multiple GPUs.

\subsection{Pre-training Experiments}
% Tab. \ref{tab:A1} provides the detail configurations for pre-training our Brainformer~\citep{zhou2024brainformers}, GLaM~\cite{du2022glam}, and Mistral~\cite{jiang2024mixtral} on \texttt{Enwik8}, \texttt{Text8}. 
Table \ref{tab:A1} provides the detailed configurations for pre-training Brainformer~\cite{zhou2024brainformers}, GLaM~\cite{du2022glam}, and Mistral~\cite{jiang2024mixtral} on \texttt{Enwik8} and \texttt{Text8}.

% \subsection{Pre-training Experiments} \label{A1}
% Tab. \ref{tab:A1} shows the implementation details used for pre-training experiments on \texttt{enwik8} and \texttt{WikiText-103} datasets. 

\begin{table}[!ht]
\centering
\setlength\tabcolsep{3.06pt}

\begin{tabular}{lccccc}
\midrule
Dataset   & Input length & Batch size & Optimizer & Lr   & \# Training Step \\ \midrule
\texttt{Enwik8}      & 512          & 48          & Adam      & 4.5e-4 & 50k       \\
\texttt{Text}      & 512          & 48          & Adam      & 4.5e-4 & 50k              \\ \midrule
\end{tabular}
\caption{Hyperparameter settings for pre-training experiments on \texttt{Enwik8} and \texttt{Text8}. }
\label{tab:A1}
\end{table}

\subsection{Finetuning Experiments}
% \noindent For finetuning experiments, we use the same model architecture as in pre-training. Tab. \ref{tab:A2} shows the detail configurations used for finetuning experiments on \texttt{SST-2}, \texttt{SST-5}, \texttt{IMDB}, and \texttt{BANKING77} datasets. 
\noindent For fine-tuning experiments, we employ the identical model architecture as in pre-training. Table \ref{tab:A2} presents the detailed configurations utilized for fine-tuning experiments on \texttt{SST-2}, \texttt{SST-5}, \texttt{IMDB}, and \texttt{BANKING77} datasets.
\begin{table}[!ht]
\centering

\setlength\tabcolsep{4.86pt}
\begin{tabular}{lccccc}
\midrule
Dataset   & Input length & Batch size & Optimizer & Lr   & \# Epochs \\ \midrule
\texttt{SST-2}     & 512          & 16         & Adam      & 1e-4 & 5         \\
\texttt{SST-5}     & 512          & 16         & Adam      & 1e-4 & 5         \\
\texttt{IMDB}      & 512          & 4          & Adam      & 1e-4 & 5         \\
\texttt{BANKING77} & 512          & 16         & Adam      & 1e-4 & 50         \\ \midrule
\end{tabular}
\caption{Detail settings for finetuning experiments on the evaluation datasets. }
\label{tab:A2}
\end{table}

\end{document}